\documentclass[lettersize,journal]{IEEEtran}
\usepackage{amsmath}
\usepackage{amsfonts}
\usepackage{amssymb}
\usepackage{bbold}
\usepackage{array}
\usepackage[caption=false,font=small,labelfont=rm,textfont=rm]{subfig}
\usepackage{url}
\usepackage{verbatim}
\usepackage{graphicx}
\usepackage{subcaption}
\usepackage{colortbl}
\usepackage{placeins}
\usepackage{booktabs}
\usepackage{balance}
\usepackage[linesnumbered,ruled,vlined]{algorithm2e}
\usepackage{threeparttable}

\begin{document}

\title{EMMa: \textbf{E}nd-Effector Stability-Oriented \textbf{M}obile \textbf{Ma}nipulation for Tracked Rescue Robots}
\author{Yifei Wang$^{1}$, Hao Zhang$^{1}$, Jidong Huang$^{1}$, Shuohang Fang$^{1}$, Haoyao Chen$^{1}$}
\maketitle

\begin{abstract}
	The autonomous operation of tracked mobile manipulators in rescue missions requires not only ensuring the reachability and safety of robot motion but also maintaining stable end-effector manipulation under diverse task demands. However, existing studies have overlooked many end-effector motion properties at both the planning and control levels. This paper presents a motion generation framework for tracked mobile manipulators to achieve stable end-effector operation in complex rescue scenarios. The framework formulates a coordinated path optimization model that couples end-effector and mobile base states and designs compact cost/constraint representations to mitigate nonlinearities and reduce computational complexity. Furthermore, an isolated control scheme with feedforward compensation and feedback regulation is developed to enable coordinated path tracking for the robot. Extensive simulated and real-world experiments on rescue scenarios demonstrate that the proposed framework consistently outperforms SOTA methods across key metrics, including task success rate and end-effector motion stability, validating its effectiveness and robustness in complex mobile manipulation tasks.
\end{abstract}

\begin{IEEEkeywords}
	Search and Rescue Robots, Mobile Manipulation, Motion and Path Planning, Disturbance Rejection
\end{IEEEkeywords}

\section{Introduction}

In disaster scenarios, autonomous rescue robots are required to demonstrate robust self-navigation in unstructured environments and the ability to perform a wide range of complex manipulation tasks. Consequently, mobile manipulators (such as quadruped-based centaur-like robots and tracked mobile manipulators) that combine strong terrain adaptability with flexible manipulation capabilities have emerged as effective solutions for performing exploration and hazardous object handling in complex rescue scenarios~\cite{delmerico2019the}.

Compared with centaur-like mobile manipulators, tracked mobile manipulators offer several distinct advantages. Their powerful drive and high operational stability enable robust mobility over rough terrains and rubble-strewn environments. Moreover, the mechanical robustness of the tracked chassis provides strong resistance to damage and external impacts. In addition, their large payload capacity and self-supporting strength allow them to carry heavier rescue equipment and execute more demanding tasks. Therefore, enabling autonomous operation of tracked mobile manipulators in disaster environments represents a pragmatic and pressing demand in rescue robotics.

\begin{figure}[h]
	\centering
	\includegraphics[width=1.0\linewidth]{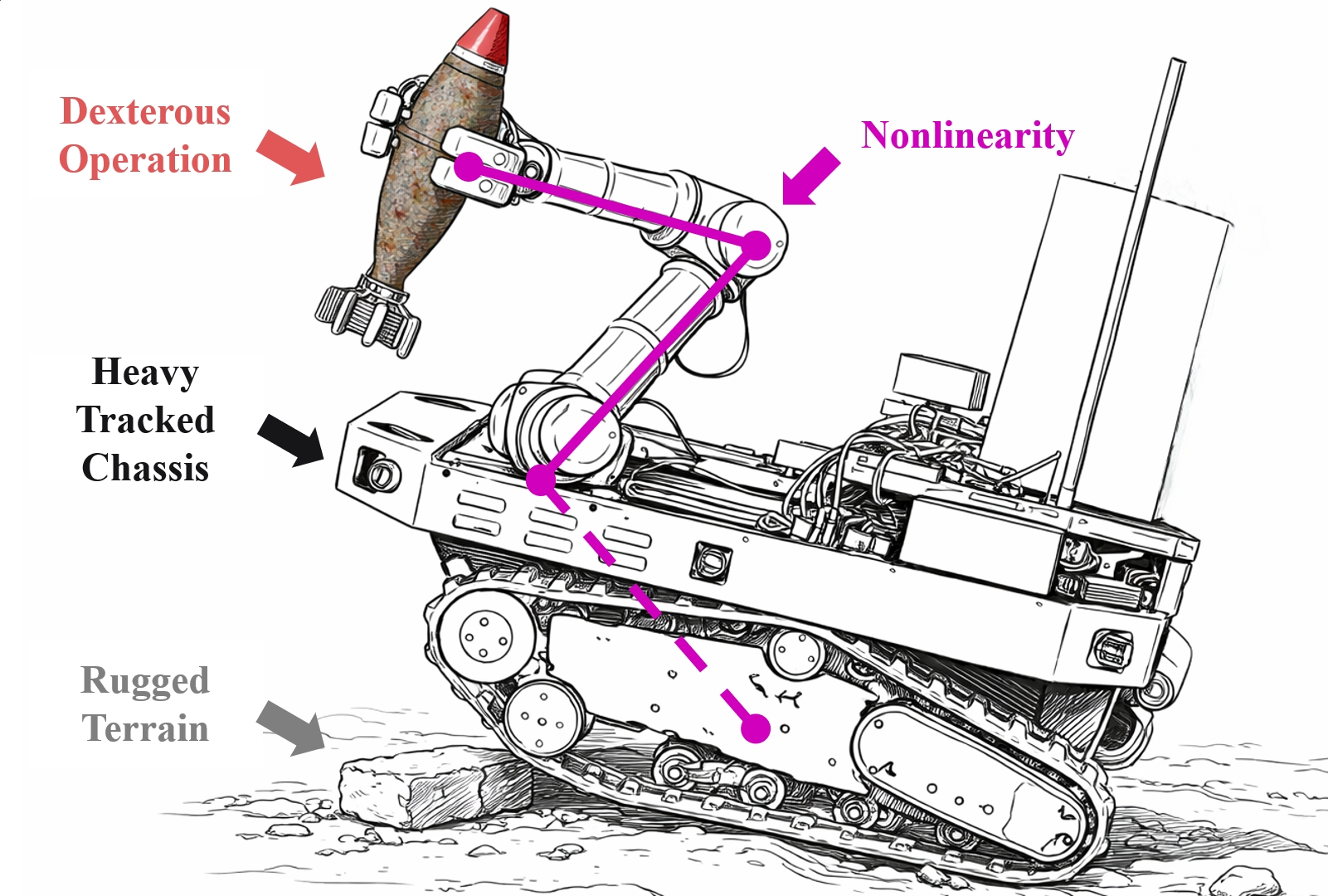}
	\captionsetup{font={small},justification=raggedright}
	\caption{The concept of mobile manipulation for explosive ordnance disposal in rescue scenarios.}
	\label{fig:00}
\end{figure}

Given the instability and hazards inherent in disaster environments, rescue robots must execute tasks with caution and avoid obstacles to prevent accidents. Meanwhile, executing fine-grained tasks such as dexterous manipulation or detailed inspection demands stable end-effector motion, which is crucial for enhancing task success and minimizing operational risks (see Fig.~\ref{fig:00}).

However, achieving stable end-effector motion for tracked mobile manipulators in rescue missions remains highly challenging. Such systems typically comprise a tracked chassis with an imprecise kinematic model and a serial manipulator with a precise one. The high degrees of freedom and strong coupling between these components introduce numerous nonlinear costs and constraints in path planning, including kinematic feasibility, obstacle avoidance, and manipulability. These nonlinearities intensify the problem's non-convexity and constraint coupling, complicating weight design and hindering real-time optimization~\cite{zhang2025an}. Although optimal control methods based on accurate whole-body dynamic or kinematic models have been widely explored for centaur-like mobile manipulators~\cite{haviland2022holistic, OCS2}, they are less effective for tracked platforms, where inaccurate robot models pose a challenge.

\IEEEpubidadjcol

Furthermore, certain rescue tasks require the robot's end-effector motion to satisfy strict hard constraints~\cite{sereinig2020review}, such as maintaining specific visual perspectives or performing precise grasping. Nevertheless, existing optimization-based approaches often struggle to satisfy hard constraints while optimizing~\cite{pankert2020pmpc}. Efficiently integrating end-effector hard constraints with the structural and motion characteristics of mobile manipulators remains an open challenge.

Finally, the large mass and inertia of the tracked chassis result in slow dynamic responses, and significant disturbances often accompany its motion over uneven terrain. These disturbances can be transmitted and even amplified through the serial manipulator structure to the end-effector, leading to substantial degradation of manipulation stability. If not properly mitigated, such disturbances may not only reduce task success rates but also pose potential safety risks~\cite{woolfrey2021pre, rigotti2018ground}. Therefore, developing a planning and control framework that ensures end-effector stability while maintaining the feasibility of whole-body motion has become an urgent challenge for tracked mobile manipulators in rescue applications.

\begin{figure*}[h]
	\centering
	\includegraphics[width=1.0\linewidth]{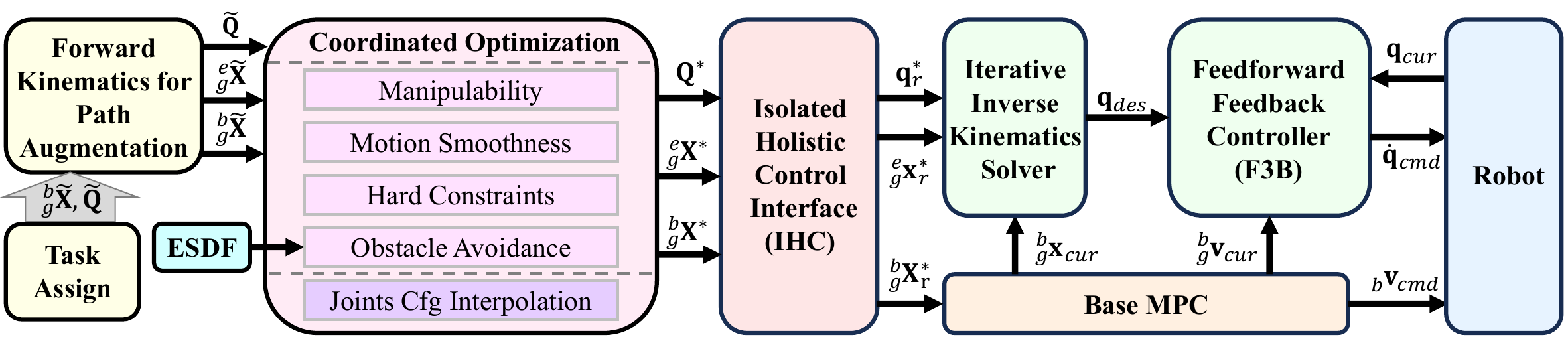}
	\captionsetup{font={small},justification=raggedright}
	\caption{General framework of EMMa.}
	\label{fig:01}
\end{figure*}

To address the aforementioned challenges, this paper presents a planning and control framework for tracked mobile manipulators aimed at stable end-effector operation. First, we formulate a coordinated optimization problem by jointly reasoning over end-effector and base states, achieving high-quality paths with improved computational efficiency. Second, by considering end-effector motion stability under chassis motion limitations, the framework ensures robust obstacle avoidance. Finally, an isolated control strategy is designed to suppress disturbances on the end-effector induced by base maneuver, thereby enhancing both response speed and end-effector control stability. The main contributions of this work are summarized as follows:
\begin{enumerate}
	\item An optimization problem is formulated for the path planning of a heavy tracked mobile manipulator. Cost functions and optimization constraints are derived by simultaneously considering manipulability, end-effector and chassis motion smoothness, and whole-body collision avoidance.
	\item An isolated holistic control solution is designed to further suppress the disturbance affection derived from the heavy chassis, caterpillar track, and rugged terrain, utilizing a feedforward feedback control scheme. 
	\item A motion planning and control framework is developed based on the proposed optimization-based planning and an isolated holistic control-based tracking controller. As we know, this is the first framework designed for stabilized end-effector motion of tracked mobile manipulators. The effectiveness is validated through extensive simulations and real-world experiments.
\end{enumerate}

\section{Related Works}

As a canonical high-DOF redundant system, the mobile manipulator has been extensively investigated in path planning under general task scenarios, including sampling-based methods~\cite{vinay2018mm, thakar2020bidi}, imitation learning methods~\cite{yan2025m2diffuser}, deep reinforcement learning (DRL) methods~\cite{li2020hrl4in, xia2021relmogen}, and others. Sampling-based approaches generally produce coarse paths and lack explicit planning of end-effector states. However, introducing appropriate state constraints during the sampling phase can partially improve path quality~\cite{pardi2020const, burget2016birrt}. Imitation learning approaches leverage expert demonstrations and network models to plan mobile manipulator motions, yet they remain limited by the difficulty of acquiring diverse expert data in complex scenarios. DRL-based methods commonly adopt a decoupled arm-base control strategy that ensures reachability and completeness in structured environments but typically overlook end-effector motion stability and manipulability.

To address the planning problem that considers the end-effector's property, one class of methods~\cite{pankert2020pmpc, dong2020catch, sleiman2021mpc} employs iterative optimization to handle nonlinear constraints in path optimization. By integrating accurate whole-body kinematic and dynamic models and utilizing sequential quadratic programming (SQP), these approaches enable joint space optimization for centaur-like mobile manipulators while explicitly accounting for end-effector costs. Another class of methods~\cite{nguyen2024plan} advocates constructing optimization problems directly in the coupled end-effector/base state space, allowing for effective optimization of end-effector-related costs or constraint satisfaction in structured environments.

Meanwhile, the pursuit of stable end-effector control on uneven terrain has motivated research aimed at enhancing mechanical design. For example, Sandy \textit{et al.}~\cite{sandy2017dyn} enhanced visual servoing performance by exploiting the fast response of linear actuators, while Mattia \textit{et al.}~\cite{rigotti2018ground} adopted hydraulic linear actuators combined with $H_{\infty}$ control and feedforward compensation to suppress terrain-induced disturbances. Moreover, some research emphasizes accurate modeling to improve control precision. Morlando \textit{et al.}~\cite{morlando2022non} developed a complete dynamic model for a centaur-like mobile manipulator to enable cross-scenario planning and control. For problems where precise modeling is challenging, data-driven disturbance-prediction~\cite{wang2025terrain} and model-estimation~\cite{woolfrey2021pre} techniques have attracted increasing attention. In addition, recent studies have shown that decoupled control strategies effectively mitigate dynamic interference arising from structural redundancy and improve the stability of end-effector control~\cite{honerkamp2021learn, picard2022multi, limerick2023motm}.

In summary, most existing mobile manipulator motion generation approaches formulate path optimization based on joint/base coupled states, with limited attention to end-effector-oriented costs and constraints. Moreover, the stability and manipulability of end-effector operation are often neglected, leading to limited adaptability and robustness in unstructured or complex environments. To address these limitations, this work incorporates specific state variables and simplified cost/constraint functions into the path optimization process, along with a tailored isolated control strategy for typical tracked mobile manipulators, enabling autonomous, stable operation in challenging rescue scenarios.

\section{Methodology}

As illustrated in Fig.~\ref{fig:01}, the proposed framework first performs forward kinematic (FK) computation based on the robot's initial base path ${_g^b\tilde{\mathbf{X}}}$ in the global frame and the initial joint configuration path $\tilde{\mathbf{Q}}$ of the manipulator, yielding the corresponding initial end-effector path ${_g^e\tilde{\mathbf{X}}}$ in the global frame. Subsequently, the framework formulates a structured representation of the optimization variable $\tilde{\mathbf{X}} = [{_g^e\tilde{\mathbf{X}}},~{_g^b\tilde{\mathbf{X}}}]$, enabling explicit modeling of multiple soft costs and hard constraints. During optimization, the manipulator's joint configuration path $\tilde{\mathbf{Q}}$ is iteratively adjusted in coordination with $\tilde{\mathbf{X}}$, jointly addressing task objectives such as end-effector manipulability, motion smoothness, and whole-body collision avoidance. Finally, the framework employs an isolated holistic control (IHC) strategy that integrates model predictive control (MPC) for the base with a feedforward-feedback (F3B) control scheme for the end-effector. This isolated design effectively suppresses disturbances transmitted from chassis motion, thereby improving the accuracy and smoothness of end-effector path tracking.

\subsection{The Coordinated End-effector/Base Path Planning for Mobile Manipulation}
\label{3.1}
\begin{figure}[h]
	\centering
	\includegraphics[width=1.0\linewidth]{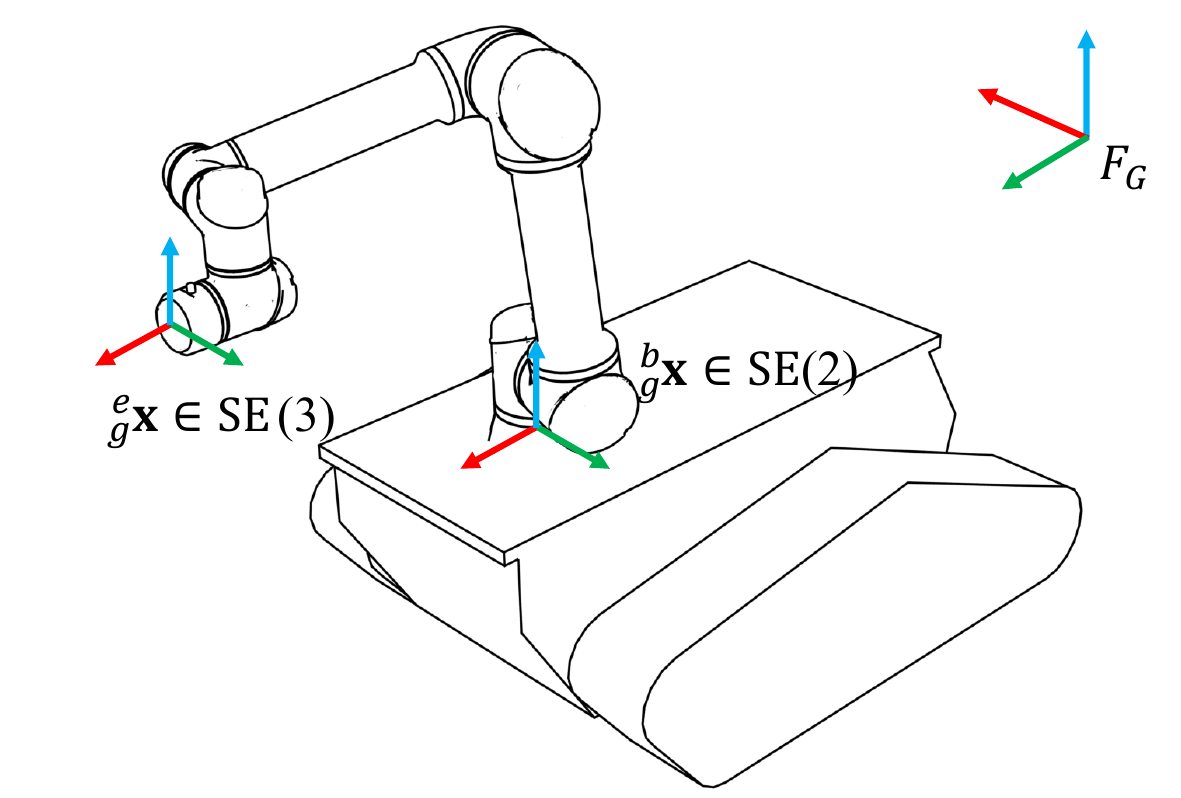}
	\captionsetup{font={small},justification=raggedright}
	\caption{The typical mechanism of a tracked mobile manipulator and its key reference frame and state definition.}
	\label{fig:02}
\end{figure}

The tracked mobile manipulator consists of a differential-drive tracked chassis and a serial manipulator, forming a typical serially connected mechanism, as shown in Fig.~\ref{fig:02}. In the conventional path optimization problem for the mobile manipulator, the end-effector state is computed from the joint and base states via forward kinematics, which renders the end-effector cost function highly nonlinear and significantly increases optimization difficulty. Although iterative methods such as SQP can handle certain classes of nonlinear optimization problems~\cite{sleiman2021mpc, sleiman2021ddp}, their convergence speed, computational cost, and optimization stability become problematic for the complex cases considered in this work. Therefore, to address the complexity of nonlinear optimization, explicitly incorporating the end-effector state into the mobile manipulation planning formulation can improve the efficiency of solving end-effector cost-related optimization problems. 

Because it is difficult to model the accurate robot kinematics in height $z$, roll $\mathcal{R}$, and pitch $\mathcal{P}$ for a tracked robot when moving over rugged terrain, a hypothesis is proposed to simplify the chassis kinematics:

\textbf{Hypothesis}: The variations in height $z$, roll $\mathcal{R}$, and pitch $\mathcal{P}$ induced by the chassis motion command $[v_x,~\omega_z]$ are bounded within finite limits when the robot operates within a spatially confined region (e.g., a certain floor of a post-disaster building).

By considering the scale of a heavy tracked chassis (ranging from sub-meter to several meters), this hypothesis is mild for a tracked mobile manipulator in typical practical applications (except when the robot begins to climb a steep stair or ramp). With this hypothesis, in the global coordinate frame $F_G$, the chassis pose is represented with only  position $x, y$ and yaw $\mathcal{Y}$, denoted as ${_g^b\mathbf{x}}:= [x_b, y_b, \mathcal{Y}_b]^\top \in \mathrm{SE(2)}$. The joint configuration of the manipulator is denoted by $\mathbf{q} \in \mathbb{R}^k$, and the pose of the end-effector is represented as ${_g^e\mathbf{x}} := [x_e, y_e, z_e, \mathcal{R}_e, \mathcal{P}_e, \mathcal{Y}_e]^\top \in \mathrm{SE(3)}$.

In the proposed framework, the end-effector pose and base pose in the global coordinate frame, $\mathbf{x}:=[{_g^e\mathbf{x}},~{_g^b\mathbf{x}}]$, are coordinately defined as the states to be optimized, while the corresponding manipulator joint configuration $\mathbf{q}$ is derived according to the planned path and kinematic requirements. The optimization problem of the proposed path planning algorithm for the mobile manipulator can be formulated as:
\begin{equation}
		\mathbf{X}^* = \mathop{\arg\min}\limits_{\mathbf{X}} \ \sum_{i=0:N\!-\!1, s \in \mathbb{S}} \mathbf{e}_s(\mathbf{x}_i, \mathcal{E}_s)^{\top} \boldsymbol{\Omega}_{s} \mathbf{e}_s(\mathbf{x}_i, \mathcal{E}_s),
	\label{eq:01}
\end{equation}
where $\mathbf{X} = (\mathbf{x}_0^{\top}, \dots , \mathbf{x}_{N\!-\!1}^{\top})^{\top} \in \mathbb{R}^{9N}$ represents the concatenated vector of $N$ waypoints along the robot path, and $i$ denotes the waypoint index. The term $\mathbf{e}_s(\cdot)$ denotes the cost function associated with each optimization objective, while $\mathcal{E}_s$ represents the external elements related to that objective. The symbol $s \in \mathbb{S}$ indexes the types of cost terms, and $\Omega_s$ denotes the corresponding weight for each cost function.

\subsubsection{Cost Function Design}

The nonlinearity can significantly increase the computational complexity and degrade the optimization stability. Therefore, the key cost terms in problem~\eqref{eq:01} will be carefully designed to eliminate the nonlinear elements according to the coordinated optimization states $[{_g^e\mathbf{x}},~{_g^b\mathbf{x}}]$. Specifically, the design focuses on three principal objectives: end-effector manipulability, end-effector motion smoothness, and tracked chassis motion smoothness. Each cost term preserves its core kinematic characteristics while employing approximate or analytical formulations to reduce computational complexity. The key cost functions are designed as follows:

\textbf{End-effector Manipulability: }During mobile manipulation, the arm must maintain sufficient manipulability~\cite{yoshikawa1985manipulability} to ensure redundancy for in-time end-effector adjustments and to enable dexterous operations when necessary. The manipulability measure $m$ of a manipulator is defined as:
\begin{equation}
	m = \sqrt{\mathrm{det}(\mathbf{J}(\mathbf{q})\mathbf{J}(\mathbf{q})^{\top})},
	\label{eq:02}
\end{equation}
where $\mathbf{J}(\mathbf{q})$ denotes the end-effector's Jacobian matrix corresponding to the joint states $\mathbf{q}$. For a given manipulator, a larger value of $m$ indicates fewer restrictions on Cartesian motion and thus higher dexterity. However, as shown in Eq.~\eqref{eq:02}, manipulability exhibits a highly nonlinear relationship with the joint states, which complicates the optimization process. Therefore, it is necessary to simplify the representation of manipulability-related costs to improve computational efficiency and convergence reliability.

\begin{figure}[h]
	\centering
	\includegraphics[width=1.0\linewidth]{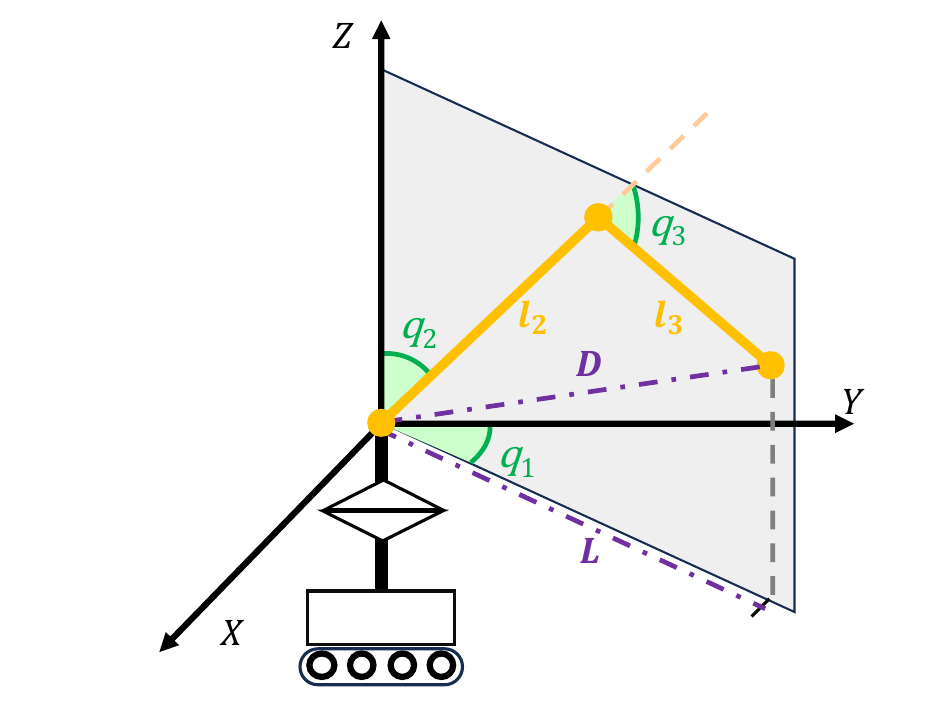}	\captionsetup{font={small},justification=raggedright}
	\caption{The typical arm span schematic description of a tracked mobile manipulator.}
	\label{fig:03}
\end{figure}

Figure~\ref{fig:03} illustrates the typical arm span of a common anthropomorphic manipulator (e.g., the UR5e) mounted on a tracked mobile base. Since the arm links near the wrist are relatively short, the manipulability of the end-effector orientation remains high and stable~\cite{yoshikawa1985manipulability}. In contrast, the end-effector's positional manipulability, denoted by $m_{\mathbf{p}}$, is mainly affected by the configurations of the shoulder lift joint $q_2$ and the elbow joint $q_3$. It can be approximated as: 
\begin{equation}
	m_{\mathbf{p}} = l_2 l_3 |(l_2 \sin(q_2) + l_3 \sin(q_2+q_3))\sin(q_3)|,
	\label{eq:03}
\end{equation}
where $l_2$ and $l_3$ denote the lengths of the upper arm and forearm, respectively.

Meanwhile, the horizontal projected Euclidean distance $L$ between the end-effector and the base on the XOY plane, as well as their direct Euclidean distance $D$, can be computed as: 
\begin{align}
	L &= |l_2\mathrm{sin}(q_2) + l_3\mathrm{sin}(q_2 + q_3)|, \label{eq:04} \\
	D &\approx \sqrt{l_2^2+ l_3^2 + 2l_2l_3\mathrm{cos}(q_3)}. \label{eq:05}
\end{align}

Combining Eqs.~\eqref{eq:03}, ~\eqref{eq:04}, and~\eqref{eq:05} , the relationship between $m_{\mathbf{p}}$, $L$, and $D$ can be derived as: 
\begin{equation}
	m_{\mathbf{p}} \approx l_2l_3 L \sqrt{1 - \Bigg(\frac{D^2 - l_2^2 + l_3^2}{2l_2l_3}\Bigg)^2}.
	\label{eq:06}
\end{equation}

Accordingly, without explicitly modeling in the manipulator joint space, the proposed framework exploits the structural characteristics of the manipulators and utilizes the relative spatial relationship between the end-effector and the base to deduce a simplified representation with less nonlinearity of the positional manipulability as: 
\begin{equation}
	\begin{aligned}
		m_{\mathbf{p}}	& \approx l_2l_3 L \sqrt{1- \Bigg(\frac{D^2 - l_2^2 + l_3^2}{2l_2l_3}\Bigg)^2}, \\
				& \approx l_2l_3 L \sin\Bigg( \frac{\pi (D+l_2-l_3)}{2l_3} \Bigg), \\
				& \approx l_2l_3 ||{_g^e\mathbf{x}^{:2}_i} {-} {_g^b\mathbf{x}^{:2}_i} ||_2 \\
				& \times \sin\Bigg( \frac{\pi \sqrt{||_g^e\mathbf{x}^{:2}_i - {_g^b\mathbf{x}^{:2}_i}||^2 + ||_g^e\mathbf{x}^{z}_i-h_b||^2}}{2l_3} \Bigg),
	\end{aligned}
	\label{eq:07}
\end{equation}
where $(\cdot)^{:2}$ denotes the first two dimensions of the optimization state, and $h_b$ denotes the height bias of the base. 

To maintain a sufficient operational margin during task execution, the reciprocal of the manipulability index $m_{\mathbf{p}}$ is adopted as the cost term:
\begin{equation}
	\mathbf{e}_m(\mathbf{x}_i, l_2, l_3) = \frac{1}{m_{\mathbf{p}}}.
	\label{eq:08}
\end{equation}
When the manipulator approaches configurations with poor manipulability, the corresponding cost increases significantly, encouraging the optimized path to preserve higher end-effector dexterity.

\textbf{End-Effector Motion Smoothness: }The motion smoothness cost is essential for enabling stable and precise mobile manipulation in complex environments. For computational tractability, the smoothness cost is decomposed into two components: the end-effector acceleration cost $\mathbf{e}_{ea}(\mathbf{x}_i, t) $ and the path curvature cost $\mathbf{e}_{e\kappa}(\mathbf{x}_i)$. 

On the one hand, the end-effector acceleration needs to be constrained to avoid excessive vibrations on the in-hand sensors or grasped objects. Given a discrete path with time interval $t$ between adjacent waypoints, the acceleration cost at waypoint $i$ can be formulated as:
\begin{equation}
	\mathbf{e}_{ea}(\mathbf{x}_i, t) = \frac{_g^e\mathbf{x}_{i\!+\!1} {-} 2_g^e\mathbf{x}_i {+} _g^e\mathbf{x}_{i\!-\!1}}{2t^2}.
	\label{eq:09}
\end{equation}

On the other hand, redundant motions should be reduced to improve execution efficiency. The directional difference between adjacent displacement vectors approximates the curvature of a discretized path, with larger differences indicating less smoothness in end-effector motion. Based on this geometric insight, the curvature smoothness cost of the end-effector path is defined as the inner product between adjacent displacement vectors, yielding:
\begin{equation}
	\mathbf{e}_{e\kappa}(\mathbf{x}_i) = 1 - ({_g^e\mathbf{x}^{\mathbf{p}}_{i\!+\!1}} - {_g^e\mathbf{x}^{\mathbf{p}}_i})\cdot({_g^e\mathbf{x}^{\mathbf{p}}_{i}} - {_g^e\mathbf{x}^{\mathbf{p}}_{i\!-\!1}}),
	\label{eq:10}
\end{equation}
where $(\cdot)^{\mathbf{p}}$ denotes the position component of the pose to be optimized. 

\textbf{Tracked Chassis Motion Smoothness: }When operating on uneven terrain, a tracked base tends to experience center-of-mass shifts and slippage during in-place rotations~\cite{zhang2024a}, and it may even be trapped by debris such as gravel, leading to mechanical faults. To alleviate these issues and improve overall maneuverability, a smoothness cost for the base motion is introduced during path planning. Since the tracked base exhibits non-holonomic constraints similar to those of differential-drive platforms, its yaw angle should remain aligned with the instantaneous direction of motion. When the yaw angle $_g^b\mathbf{x}^\mathcal{Y}_i$ at waypoint $i$ aligns with the displacement vector $(\Delta _g^b\mathbf{x}^x_{i\!-\!1}, \Delta _g^b\mathbf{x}^y_{i\!-\!1})$ between waypoints $i$ and $i\!-\!1$, the chassis motion becomes smoother and more consistent. The corresponding cost is formulated as:
\begin{equation}
	\mathbf{e}_{bk}(\mathbf{x}_i) {=} |_g^b\mathbf{x}^\mathcal{Y}_i - \text{atan}( \frac{\Delta _g^b\mathbf{x}^y_{i\!-\!1}}{\Delta _g^b\mathbf{x}^x_{i\!-\!1}})|.
	\label{eq:11}
\end{equation}

\subsubsection{Hard Constraint Design}

Path optimization for a mobile manipulator will exhibit strong coupling between its costs and constraints. However, to accelerate the optimization procedure in this work, the coordinated state to be optimized is decomposed into the base pose on $\mathrm{SE}(2)$ and the end-effector pose on $\mathrm{SE}(3)$ with hard constraints ensuring feasibility, formulated as:
\begin{equation}
		\mathbf{X}^*= \mathop{\arg\min}\limits_{\mathbf{X}}  \sum_{i=0:N\!-\!1, s \in \mathbb{S}} \mathbf{e}_s({_g^e\mathbf{x}_i},\! {_g^b\mathbf{x}_i},\! \mathcal{E}_s)^{\top} \boldsymbol{\Omega}_{s} \mathbf{e}_s({_g^e\mathbf{x}_i},\! {_g^b\mathbf{x}_i},\! \mathcal{E}_s).
	\label{eq:a}
\end{equation}
 The decomposed formulation enables a unified treatment of complex, coupled costs while ensuring the satisfaction of critical kinematic constraints. By decomposing the state variables, the algorithm avoids redundant computations and enhances the sparsity of the Hessian matrix during the optimization~\cite{dellaert2006square}, thereby improving computational efficiency.

In mobile manipulation, two typical types of hard constraints are commonly imposed on the end-effector: partial-pose constraints and complete-pose constraints. On the one hand, considering the coordinated optimization of the base and the end-effector, the end-effector position must remain within a bounded workspace relative to the base. Specifically, the distance between the end-effector and the mobile base, defined as $D = \sqrt{|| {_g^e\mathbf{x}_i^{:2}} - {_g^b\mathbf{x}_i^{:2}} ||^2 + || _g^e\mathbf{x}_i^z-h_b ||^2}$, should not exceed the workspace threshold $\epsilon_{ew}$. The corresponding \textbf{End-Effector Workspace Constraint} is formulated as:
\begin{equation}
	\mathbf{h}_{ew}(\mathbf{x}_i) = || {_g^e\mathbf{x}_i^{:2}} - {_g^b\mathbf{x}_i^{:2}} ||^2 + || _g^e\mathbf{x}_i^z - h_b||^2 - \epsilon_{ew}^2 \leq 0.
	\label{eq:12}
\end{equation}

On the other hand, for many mobile manipulation tasks (e.g., grasping during locomotion), the end-effector is required to maintain a fixed desired pose $_g^e\mathbf{x}_{des}$, while the base continues to move, thereby improving task efficiency. In this case, within a given time window, the end-effector pose must exactly match the desired pose. The corresponding \textbf{End-Effector Fixed-Pose Constraint} can be expressed as:
\begin{equation}
	\mathbf{h}_{en}(\mathbf{x}_i) = || {_g^e\mathbf{x}_i} - {_g^e\mathbf{x}_{des}} || = \mathbf{0}, i \in (s, ..., t). 
	\label{eq:13}
\end{equation}

For the fixed-pose constraint in Eq.~\eqref{eq:13}, the proposed method directly fixes the end-effector waypoints corresponding to the constrained indices in the optimization problem~\eqref{eq:a}, following the principle in~\cite{kummerle2011g2o}. For the workspace constraint in Eq.~\eqref{eq:12}, the inequality condition is reformulated as a penalty term augmented with a Lagrange multiplier, expressed as:
\begin{equation}
	\mathbf{e}_{ew}(\mathbf{x}_i) = \lambda(|| {_g^e\mathbf{x}_i^{:2}} - {_g^b\mathbf{x}_i^{:2}} ||^2 + || _g^e\mathbf{x}_i^z-h_b||^2 - \epsilon_{ew}^2),
	\label{eq:14}
\end{equation}
where $\lambda$ denotes the Lagrange multiplier, which is updated prior to each optimization iteration as:
\begin{equation}
	\lambda =
	\begin{cases} 
	10^4,  & || {_g^e\mathbf{x}_i^{:2}} - {_g^b\mathbf{x}_i^{:2}} ||^2 + || _g^e\mathbf{x}_i^z-h_b ||^2 - \epsilon_{ew}^2 > 0, \\
	0, & \mathrm{else}.
	\end{cases}
	\label{eq:15}
\end{equation}
Integrating the end-effector's hard constraints into an efficient unconstrained optimization framework is able to balance the computational efficiency of optimization with the rigor of strict constraint enforcement.

\subsection{Holistic Obstacle Avoidance}

\begin{figure}[h]
	\centering
	\includegraphics[width=0.49\textwidth]{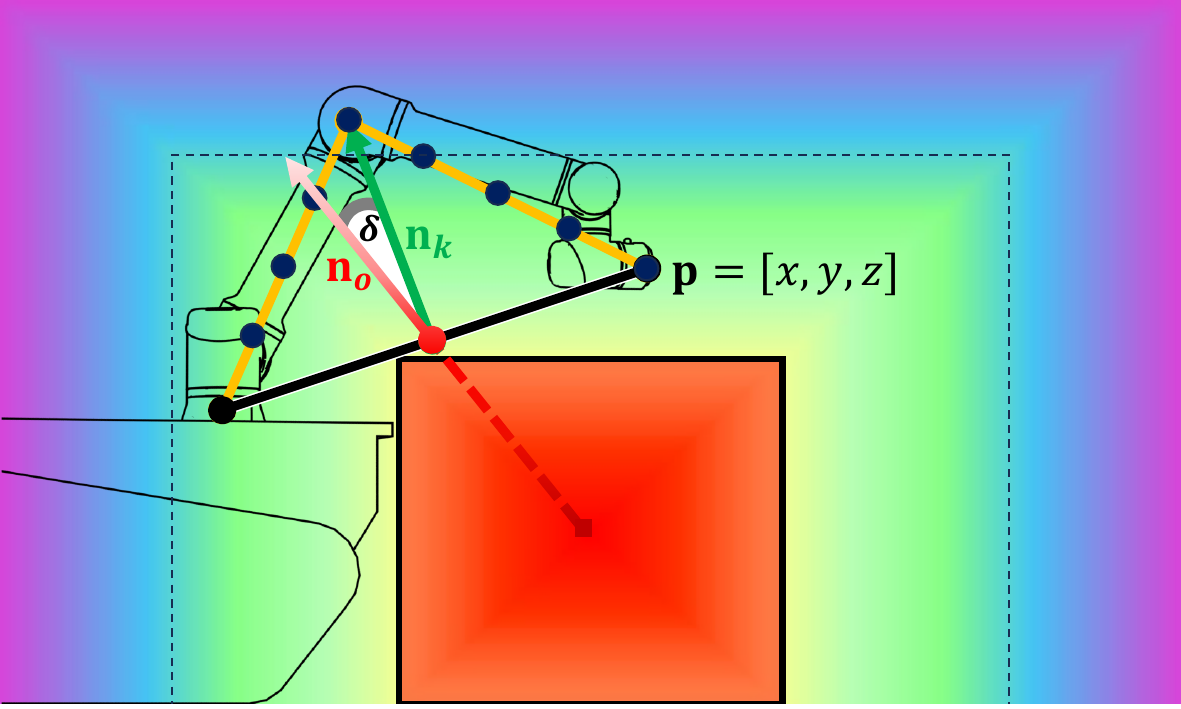}
	\captionsetup{font={small},justification=raggedright}
	\caption{Manipulator configuration selection near obstacles. In the color-coded ESDF map, red squares indicate obstacles in the environment. The green arrow $\mathbf{n}_k$ denotes the midline vector pointing toward the elbow, and the red arrow $\mathbf{n}_o$ represents the ESDF gradient direction at the midpoint of the line from end-effector to base.}
	\label{fig:04}
\end{figure}

For mobile manipulators with a serial mechanism, the main environmental collision envelopes are associated with the end-effector, the main arm links (upper and forearm segments), and the mobile base. To evaluate potential obstacle risks along the motion path, a set of 3D points $\mathbf{p}:=[x, y, z]$ (a total of $m$ points) is uniformly sampled along the centerline through the robot's base, elbow, and end-effector, as illustrated in Fig.~\ref{fig:04}. The Euclidean Signed Distance Field (ESDF) values at these sampled points are then used to assess the distance between the robot and surrounding obstacles.  

Considering the joint angle limits, for a given end-effector/base poses $[{_g^e\mathbf{x}_i}, {_g^b\mathbf{x}_i}]$, the anthropomorphic manipulator generally admits a finite set of inverse kinematic (IK) solutions $\mathbf{Q}_i = \{ \mathbf{q}_1, \mathbf{q}_2, \dots, \mathbf{q}_l \}$, each corresponding to a distinct manipulator configuration~\cite{diankov_thesis}. For a specific joint configuration $\mathbf{q}_j$, the associated sampling point set can be defined as $\mathbf{P}_j = \{ \mathbf{p}_1, \mathbf{p}_2, \dots, \mathbf{p}_m \}$. 

The collision-checking strategy evaluates all IK solutions at each waypoint using the finite set of sampled points to identify feasible whole-body configurations. Path optimization is triggered only when no collision-safe IK solution exists, while unnecessary checks are skipped when all IK solutions are safe. Otherwise, the algorithm selects the configuration that maximizes manipulability by aligning the arm posture with the outline of the obstacle. The collision-checking procedure is detailed in Algorithm~\ref{alg:01}.

\begin{algorithm}[h]
\captionsetup{font={small},justification=raggedright}
\caption{Holistic Obstacle Avoidance}
\label{alg:01}
\SetKwInOut{Input}{Input}
\SetKwInOut{Output}{Output}
\Input{$\tilde{\mathbf{X}}, \tilde{\mathbf{Q}}$; $\mathcal{SDF}$; $\epsilon_{ad}$, $\epsilon_{bd}$, $\epsilon_{ed}$;}
\For {$\mathbf{each} \ _g\mathbf{x}_i = [{_g^e\mathbf{x}_i},~{_g^b\mathbf{x}_i}] \in \tilde{\mathbf{X}}$}{
    $\mathbf{Q}_i = \{ \mathbf{q}_1, \dots, \mathbf{q}_l \} = \texttt{IKFastSolver}(_g\mathbf{x}_i )$\;
    \For {$\mathbf{each} \ \mathbf{q}_j \in \mathbf{Q}_i$}{
        $\mathbf{P}_j = \{ \mathbf{p}_1, \dots, \mathbf{p}_m \} = \texttt{Sample}(\mathbf{q}_j)$\;
        \If{$\forall \mathbf{p}_k \in \mathbf{P}_j, \mathcal{SDF}_{dist}(\mathbf{p}_k) > \epsilon_{ad}$  \\
        \hspace*{1.1em} $\mathbf{and} ~ \mathcal{SDF}_{dist}(_g^e\mathbf{x}_i) > \epsilon_{ed}$ \\
        \hspace*{1.1em} $\mathbf{and} ~ \mathcal{SDF}_{dist}(_g^b\mathbf{x}_i) > \epsilon_{bd}$}{
            $\mathbf{Q}_i^s \leftarrow \mathbf{Q}_i^s .\texttt{append}(\mathbf{q}_j)$\;
        }
    }
    \If{$\mathbf{size}(\mathbf{Q}_i^s) == 0$}{
        \texttt{setPathOptimization$(_g\mathbf{x}_i)$}\;
        \Return\;
    }
    \If{$\mathbf{size}(\mathbf{Q}_i^s) == l$}{
        \Return\;
    }
    $\mathbf{q}_{i,new} =\texttt{getBestCfg}(\mathbf{Q}_i^s, \mathcal{SDF})$\;
    $\tilde{\mathbf{Q}}.\texttt{update}(\mathbf{q}_{i,new}, i)$\;
}
\end{algorithm}

For each path waypoint $_g\mathbf{x}_i = [{_g^e\mathbf{x}_i}, {_g^b\mathbf{x}_i}]$, a corresponding inverse solution $\mathbf{q}_j$ is regarded as feasible and added to the feasible solution set $\mathbf{Q}_i^s$ if all sampled points in $\mathbf{P}_j$ have ESDF distances larger than the arm safety threshold $\epsilon_{ad}$, and meanwhile the waypoint $_g\mathbf{x}_i$ satisfy the end-effector and mobile base safe obstacle clearance $\epsilon_{ed}$ and $\epsilon_{bd}$, respectively (see Algorithm~\ref{alg:01}, Lines 4-8).

If no feasible inverse solution exists at the waypoint (i.e., the size of the feasible inverse solution set $\mathbf{Q}_i^s$ equals $0$), it indicates that the corresponding waypoint violates the collision-avoidance constraints. In this case, the waypoint $_g\mathbf{x}_i$ is marked for path optimization and adjustment (see Algorithm~\ref{alg:01}, Lines 9-11). During the optimization, to ensure a safe clearance between the robot and surrounding obstacles $O$, the collision-avoidance cost is composed of two components: (a) the \textbf{Positional Distance Cost} $\mathbf{e}_{op}$, defined as the reciprocal of the ESDF distance, which directly penalizes waypoints approaching the obstacle; and (b) the \textbf{Velocity Direction Cost} $\mathbf{e}_{ov}$, which encourages the robot's motion direction to deviate from the gradient of the obstacle distance field, thereby preventing movement toward obstacles. To accommodate the proposed optimization problem~\eqref{eq:a}, these components are formulated as follows:
\begin{align}
	\mathbf{e}_{op}(\mathbf{x}_i, \mathcal{SDF})
		&= \frac{1}{d(_g\mathbf{x}^{\mathbf{p}}_{i}, O)}
		= \frac{1}{\mathcal{SDF}_{dist}(_g\mathbf{x}^{\mathbf{p}}_{i})}, \label{eq:16}\\
	\mathbf{e}_{ov}(\mathbf{x}_i, \mathcal{SDF})
		&= 1- ({_g\mathbf{x}^{\mathbf{p}}_{i\!+\!1}} - {_g\mathbf{x}^{\mathbf{p}}_i})
		\cdot \mathcal{SDF}_{grad}(_g\mathbf{x}^{\mathbf{p}}_{i}), \label{eq:17}
\end{align}
where $\mathcal{SDF}_{dist}(_g\mathbf{x}^{\mathbf{p}}_{i})$ and $\mathcal{SDF}_{grad}(_g\mathbf{x}^{\mathbf{p}}_{i})$ denote the ESDF distance and its gradient at the end-effector/base position $_g\mathbf{x}^{\mathbf{p}}_{i}$ in the global frame, respectively.

If all inverse solutions in $\mathbf{Q}_i$ are feasible (i.e., the size of the feasible inverse solution set $\mathbf{Q}_i^s$ equals $l$), collision avoidance is omitted at this waypoint to reduce computational overhead, since no nearby obstacles effect the mobile manipulator (see Algorithm~\ref{alg:01}, Lines 12-13).

Otherwise, when only a subset of inverse solutions is feasible, the algorithm selects the most suitable configuration from $\mathbf{Q}_i^s$. As illustrated in Fig.~\ref{fig:04}, each inverse solution $\mathbf{q}_j$ can be geometrically represented by a triangle formed by the mobile base, elbow, and end-effector. To maximize the manipulator's operational capability while maintaining obstacle clearance, the algorithm selects the feasible inverse solution whose elbow midline vector $\mathbf{n}_k$ has the smallest directional difference $\delta$ with the ESDF gradient vector $\mathbf{n}_o$. The corresponding inverse solution $\mathbf{q}_{i, new}$ is then updated to the corresponding joint path $\tilde{\mathbf{Q}}$ (see Algorithm~\ref{alg:01}, Lines 14-15).

\subsection{Joint Configuration Interpolation for Motion Stability}

During the coordinated optimization of the end-effector and base states, the arm configuration corresponding to adjacent waypoints may undergo significant changes due to obstacle avoidance and other requirements. The planner must therefore ensure not only the smoothness and stability of the end-effector's state but also prevent abrupt changes in joint configurations. To address this issue, the proposed framework adopts a joint state interpolation method that accounts for both end-effector stability and base motion constraints, without explicitly optimizing the manipulator joint variables.

\begin{figure}[h]
	\centering
	\includegraphics[width=0.49\textwidth]{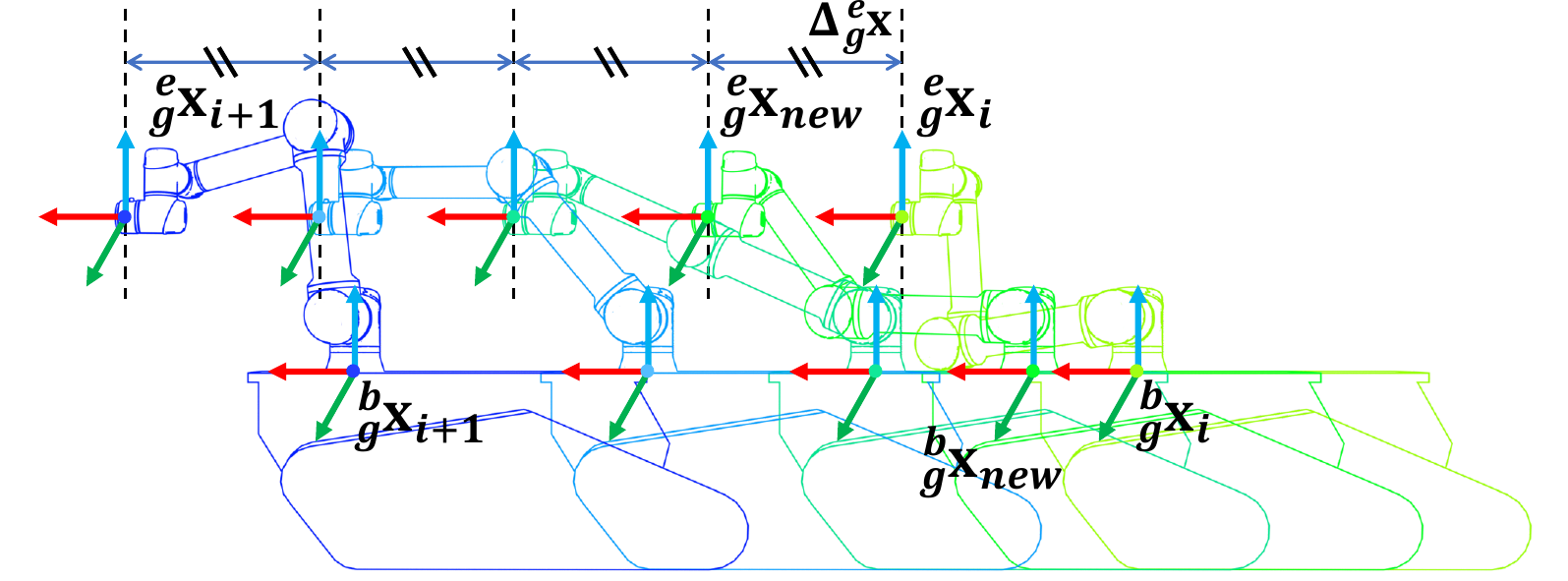}
	\captionsetup{font={small},justification=raggedright}
	\caption{Joint configuration and base path interpolation considering end-effector stability.}
	\label{fig:05}
\end{figure}

As illustrated in Fig.~\ref{fig:05}, this approach smooths whole-body motion by locally refining path segments where adjacent arm joint configurations change abruptly, enforcing end-effector stability while constraining joint-induced base motion to the planar XOY manifold through null-space projection. The detailed interpolation procedure is summarized in Algorithm~\ref{alg:02}.

\begin{algorithm} 
\captionsetup{font={small},justification=raggedright}
\caption{Joint Configuration Interpolation}
\label{alg:02}
\SetKwInOut{Input}{Input}
\SetKwInOut{Output}{Output}
\Input{$\tilde{\mathbf{X}}, \tilde{\mathbf{Q}}$; threshold $\delta_{J}$; $n\!=\!7$; $a\!=\!0$;}
\Output{Inserted $\tilde{\mathbf{X}}, \tilde{\mathbf{Q}}$;}
\For{$\mathbf{each} ~ (_g\mathbf{x}_i,~_g\mathbf{x}_{i\!+\!1})  \in \tilde{\mathbf{X}},~ (\mathbf{q}_i,~\mathbf{q}_{i\!+\!1})  \in \tilde{\mathbf{Q}}$}{
    \If{$|\|\mathbf{J}_{b}^{e}(\mathbf{q}_i)\|_{\mathrm{F}} - \|\mathbf{J}_{b}^{e}(\mathbf{q}_{i\!+\!1})\|_{\mathrm{F}}| > \delta_{J}$}{
        $\Delta \tilde{\mathbf{q}} = (\mathbf{q}_{i\!+\!1}- \mathbf{q}_i)/n$\;
        $\Delta _g^e\mathbf{x} = (_g^e\mathbf{x}_{i\!+\!1} - {_g^e\mathbf{x}_{i}})/n$; \% In SE(3) space\
        \While{$a \leq n$}{
            $_g^e\mathbf{x}_{new} = {_g^e\mathbf{x}_{i+a}} \oplus \Delta _g^e\mathbf{x}$\;
            $_\text{s}\mathbf{J} = \mathbf{J}_{g}^{b}(\mathbf{q}_{i+a})_{[2:5]}$\;
            $\mathbf{P}_{\mathcal{N}(\mathbf{J})} = \mathbf{I} - {_\text{s}}\mathbf{J}^{\top}(_\text{s}\mathbf{J}{_\text{s}}\mathbf{J}^{\top})^{-1} {_\text{s}}\mathbf{J}$\;
            $\Delta \mathbf{q}^* = \mathbf{P}_{\mathcal{N}(\mathbf{J})} \cdot \Delta \tilde{\mathbf{q}}$\;
            $\mathbf{q}_{new} = \mathbf{q}_{i+a} + \Delta \mathbf{q}^*$\;
            $_g^b\mathbf{x}_{new} = \texttt{FKSolver}(\mathbf{q}_{new}) \oplus {_g^e\mathbf{x}_{new}} $\;
            $\tilde{\mathbf{X}}.\texttt{insert}([_g^e\mathbf{x}_{new},~{_g^b\mathbf{x}_{new}}], i, a)$\;
            $\tilde{\mathbf{Q}}.\texttt{insert}(\mathbf{q}_{new}, i, a)$\;
            $a\!+\!+$\;
        }
    }
}
\Return $\tilde{\mathbf{X}}, \tilde{\mathbf{Q}}$\;
\end{algorithm}

The algorithm first compares the adjacent manipulator configurations $(\mathbf{q}_i,~\mathbf{q}_{i\!+\!1})$ associated with their consecutive waypoints $(_g\mathbf{x}_i,~ _g\mathbf{x}_{i\!+\!1})$. If the difference in the Frobenius norm $\|\cdot\|_{\mathrm{F}}$ of their end-effector Jacobian matrices $\mathbf{J}_{b}^{e}(\mathbf{q}_i)$ and $\mathbf{J}_{b}^{e}(\mathbf{q}_{i\!+\!1})$ with respect to the base frame exceeds a predefined threshold $\delta_{J}$, the corresponding joint configurations are regarded as significantly different (see Algorithm~\ref{alg:02}, Line 2). Considering the requirement of end-effector stability, the algorithm then performs $n$ uniform interpolations between $(_g^e\mathbf{x}_i,~{_g^e\mathbf{x}_{i\!+\!1}})$ in SE(3) space (see Algorithm~\ref{alg:02}, Lines 3-6), and subsequently conducts joint space interpolation under the constraint of maintaining the interpolated end-effector poses $_g^e\mathbf{x}_{new}$ unchanged.

According to the hypothesis in Section~\ref{3.1}, the base state along the path is constrained within the manifold of the XOY plane. Consequently, the base motion tendency during the interpolation should also be constrained within this space:
\begin{equation}
	\mathbf{J}_{g}^{b}(\mathbf{q})_{[2:5]} \cdot \Delta \mathbf{q^*} = \mathbf{0}^{3\times 1},
	\label{eq:18}
\end{equation}
where $\mathbf{J}_{g}^{b}(\mathbf{q})$ denotes the mobile base's Jacobian matrix with respect to the global frame when the end-effector's pose is fixed; $\mathbf{J}_{g}^{b}(\mathbf{q})_{[2:5]}$ denotes the sub-matrix formed by the 2nd to 5th rows of $\mathbf{J}_{g}^{b}(\mathbf{q})$. The constraint \eqref{eq:18} indicates that, when the end-effector pose is held constant, the interpolated joint motions $\Delta \mathbf{q^*}$ should induce no motion in the base's height $z$, roll $\mathcal{R}$, or pitch $\mathcal{P}$, thereby ensuring end-effector stability while respecting base motion constraints.

For the uniformly interpolated joint increments $\Delta \tilde{\mathbf{q}}$ (see Algorithm~\ref{alg:02}, Line 3), a null-space projection method is employed to ensure that the final joint update $\Delta \mathbf{q^*}$ that satisfies the equality constraint in Eq.~\eqref{eq:18} (see Algorithm~\ref{alg:02}, Lines 7-9), with:
\begin{equation}
	\Delta \mathbf{q^*} = \mathbf{P}_{\mathcal{N}(\mathbf{J})} \cdot \Delta \tilde{\mathbf{q}}.
	\label{eq:20}
\end{equation}

And the corresponding projection matrix $\mathbf{P}_{\mathcal{N}(\mathbf{J})}$ is computed with $\mathbf{J}_{g}^{b}(\mathbf{q})_{[2:5]} \in \mathbb{R}^{3\times k}$ denoted as ${_\text{s}}\mathbf{J}$: 
\begin{equation}
 \mathbf{P}_{\mathcal{N}(\mathbf{J})} = \mathbf{I} - {_\text{s}}\mathbf{J}^{\top}(_\text{s}\mathbf{J}_\text{s}\mathbf{J}^{\top})^{-1}{_\text{s}}\mathbf{J}.
	\label{eq:21}
\end{equation}
 
Subsequently, given the interpolated end-effector poses $_g^e\mathbf{x}_{new}$ and the corresponding joint configurations $\mathbf{q}_{new}$, the framework employs forward kinematics to compute the associated base pose $_g^b\mathbf{x}_{new}$, thereby generating a new coordinated waypoint (see Algorithm~\ref{alg:02}, Lines 10-13). This new waypoint and arm configuration serves as a reference in subsequent iterations, allowing continuous interpolation updates of the paths until a smooth transition across the segment is achieved.

\subsection{Isolated Holistic Control for Mobile Manipulation}

In unstructured environments with rough terrain, the locomotion of a tracked chassis often induces disturbances or oscillations that propagate along the arm, posing significant challenges for precise end-effector control. To mitigate such coupled perturbations, this work proposes an isolated holistic control strategy designed to suppress the influence of base disturbances on the end-effector.

As illustrated in Fig.~\ref{fig:01}, the mobile base executes the path-tracking task using a model predictive controller to follow the optimized base path ${_g^b\mathbf{X}^*}$. Building upon this, a feedforward-feedback control scheme is proposed to stabilize the end-effector state. First, by combining the reference end-effector pose ${_g^e\mathbf{x}^{*}_{r}}$ (from the optimized end-effector path ${_g^e\mathbf{X}^{*}}$) with the current base state ${_g^b\mathbf{x}_{cur}}$, the joint reference configuration $\mathbf{q}^{*}_{r}$ (from the corresponding arm joints path $\mathbf{Q}^{*}$) is used as the initial condition for iterative inverse kinematics, yielding the desired joint configuration $\mathbf{q}_{des}$. Subsequently, a feedback correction term based on the joint error $\mathbf{q}_{err} = \mathbf{q}_{des} - \mathbf{q}_{cur}$, is designed to ensure stable end-effector pose tracking. Then, by incorporating the mobile base motion, the coupling effect between the base and end-effector is estimated. As both the joint configuration and base pose undergo only slight instantaneous variations, the end-effector velocity ${_g^e\mathbf{v}_{ind}}$ induced by the base velocity ${_g^b\mathbf{v}_{cur}}$ in the global frame can be directly derived through forward kinematics. The induced end-effector velocity ${_b^e\mathbf{v}_{ind}}$ in the base frame is then calculated by:
\begin{equation}
    {_b^e\mathbf{v}_{ind}} =
    \begin{bmatrix}
        {_b^e\mathbf{v}_{ind}^l} \\
        {_b^e\mathbf{v}_{ind}^a}
    \end{bmatrix} =
    \begin{bmatrix}
        {_g^b\mathbf{v}_{cur}^{l}} + {_g^b\mathbf{v}_{cur}^{a}} {\times} {_b^e\mathbf{x}_{cur}}\\
        {_g^b\mathbf{v}_{cur}^{a}}
    \end{bmatrix},
    \label{eq:22}
\end{equation}
where $(\cdot)^l$ and $(\cdot)^a$ are the linear and angular components of the velocity, respectively.

Subsequently, the induced chassis-caused end-effector equivalent velocity ${_b^e\mathbf{v}_{ind}}$ is mapped into the joint space through the Jacobian matrix, yielding the joint space motion feedforward term:
\begin{equation}
	\dot{\mathbf{q}}_{ind} = {\mathbf{J}_b^e(\mathbf{q}_{cur})^{-1}} \cdot {_b^e\mathbf{v}_{ind}}.
	\label{eq:23}
\end{equation}
Accordingly, the F3B control policy can be formulated as the superposition of a feedback regulation policy and a feedforward compensation policy: 
\begin{equation}
    \pi_{\mathrm{f3b}} = \pi_{\mathrm{ff}}|_{\dot{\mathbf{q}}_{ind}} + \pi_{\mathrm{fb}}|_{\mathbf{q}_{err}}.
	\label{eq:24}
\end{equation}
The feedback component $\mathbf{u}_{\mathrm{fb}}(t)$ adopts a typical formulation as:
\begin{equation}
\mathbf{u}_{\mathrm{fb}}(t)
=
\pi_{\mathrm{fb}}\!\left(
\mathbf{q}_{err}(t),\,
\mathcal{H}_{\mathbf{e}}(t)
\right)
,
\quad
\pi_{\mathrm{fb}}:\; \mathcal{E} \times \mathcal{H}_{\mathcal{E}} \rightarrow \mathcal{U}
,
\label{eq:25}
\end{equation}
which regulates the joint space tracking error $\mathbf{q}_{err}(t)$. With error history $\mathcal{H}_{\mathbf{e}}(t)$, this term shapes stable closed-loop error dynamics in discrete time, driving the tracking error toward zero. It also suppresses steady-state biases caused by external disturbances.

Built upon this error-driven stability design, a feedforward component $\mathbf{u}_{\mathrm{ff}}(t)$ is introduced to further improve the transient response:
\begin{equation}
\mathbf{u}_{\mathrm{ff}}(t)
=
\pi_{\mathrm{ff}}(\dot{\mathbf{q}}_{ind}(t), \mathcal{D}^{n}_{\mathbf{y}}(t))
,
\quad
\pi_{\mathrm{ff}}:\; \mathcal{Y} \times \mathcal{D}^{n}_{\mathcal{Y}} \rightarrow \mathcal{U}
.
\label{eq:26}
\end{equation}
This term provides anticipatory compensation for predictable motion-induced components using the derivatives $\mathcal{D}^{n}_{\mathbf{y}}(t)$ up to $n$ orders, reshaping the error dynamics without compromising closed-loop stability and thereby reducing the burden on feedback regulation.

The proposed IHC treats base motion as a predictable external disturbance and compensates for its effect at the joint level through kinematic propagation, thereby achieving end-effector/base decoupling at the control architecture level. Without explicitly modeling whole-body dynamics, this feedforward-feedback design ensures effective disturbance rejection and operational stability, preserving bounded-input-bounded-error behavior, and enabling practical asymptotic convergence of the tracking error.

\section{Simulation}
\label{simulation}

\begin{figure}[h]
	\centering
	\includegraphics[width=0.49\textwidth]{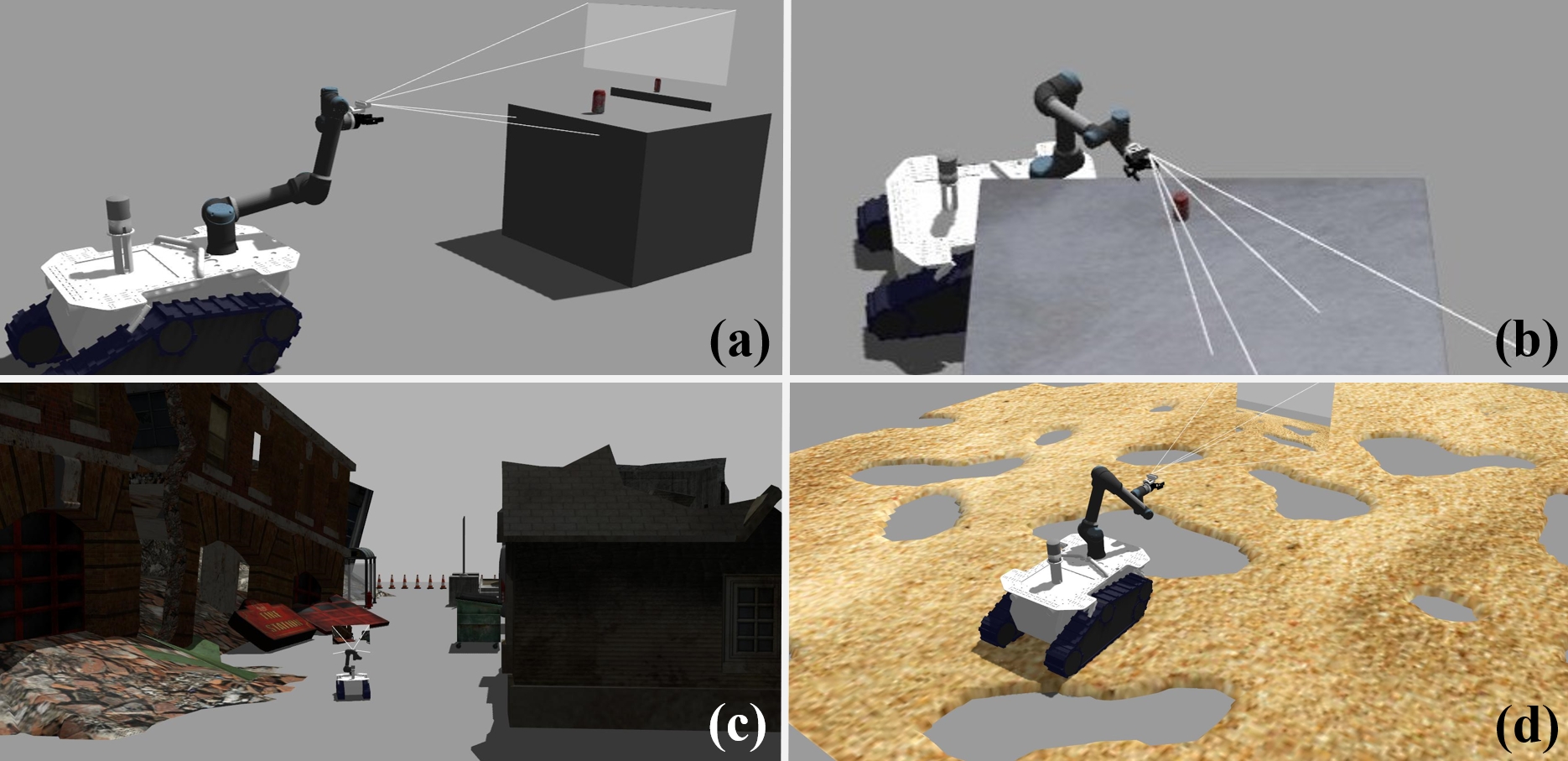}
	\captionsetup{font={small},justification=raggedright}
	\caption{Four typical simulated rescue scenarios with distinct end-effector operation requirements: (a) obstacle-aware grasping with configuration adjustment; (b) grasping during locomotion; (c) environment inspection during locomotion; and (d) transportation of a force-sensitive payload over rough terrain.}
	\label{fig:06}
\end{figure}

We reproduced several optimization-based planning and control methods for mobile manipulators and compared them with the proposed framework across multiple simulated rescue scenarios. Following the RTB~\cite{corke2021not} and HRQP controllers~\cite{haviland2022holistic}, we implemented the mobile manipulation framework proposed by Limerick \textit{et al.}~\cite{limerick2023motm,limerick2024reactive}, hereafter referred to as \textbf{ReDyn}. For MPC-based integrated planning-control methods that leverage whole-body kinematic models of mobile manipulators, we developed a combined scheme of Go~Fetch!~\cite{zimmermann2021gofetch} and P-MPC~\cite{pankert2020pmpc} on top of the OCS2 framework~\cite{OCS2}, hereafter denoted as \textbf{GP}. In the simulations, all methods were deployed in four representative rescue scenarios, as illustrated in Fig.~\ref{fig:06}, and were systematically evaluated under identical environmental and task configurations. The representative parameters used in our framework are summarized in Table~\ref{table:01}.

\begin{table}[h]
    \centering
    \captionsetup{font={small},justification=raggedright}
    \caption{Typical parameter settings}
	\small
	\setlength{\tabcolsep}{3pt}
    \begin{tabular}{lll}
    \toprule
    \textbf{Parameters} & \textbf{Description} & \textbf{Value} \\
    \midrule
    $\epsilon_{ed},\epsilon_{bd}, \epsilon_{ad}$ & clearances of EE/Base/Arm & 0.05/0.2/0.1~m \\
    $\epsilon_{ew}$ & arm workspace limitation & 0.85~m \\
    $\delta_{J}$ & Frobenius norm threshold & 3.0 \\
    $\Omega_{ea},\Omega_{e\kappa}, \Omega_{ew}$ & weight of the EE costs & $10^2,10^2,10^3$ \\
    $\Omega_{bk}$ & weight of the Base cost & $10^2$ \\
    $\Omega_{m}$ & weight of the Manipulability& $50$ \\
    $\Omega_{op}, \Omega_{ov}$ & weight of the collision avoidance & $10^2, 10^2$ \\
    $K_p, K_d, N_0$ & PID/DFF gains of F3B & $3.0, 0.1, -0.3$ \\
    \bottomrule
    \end{tabular}
    \label{table:01}
\end{table}

\subsection{Obstacle-Aware Grasping}

\begin{figure}[h]
	\centering
	\includegraphics[width=0.49\textwidth]{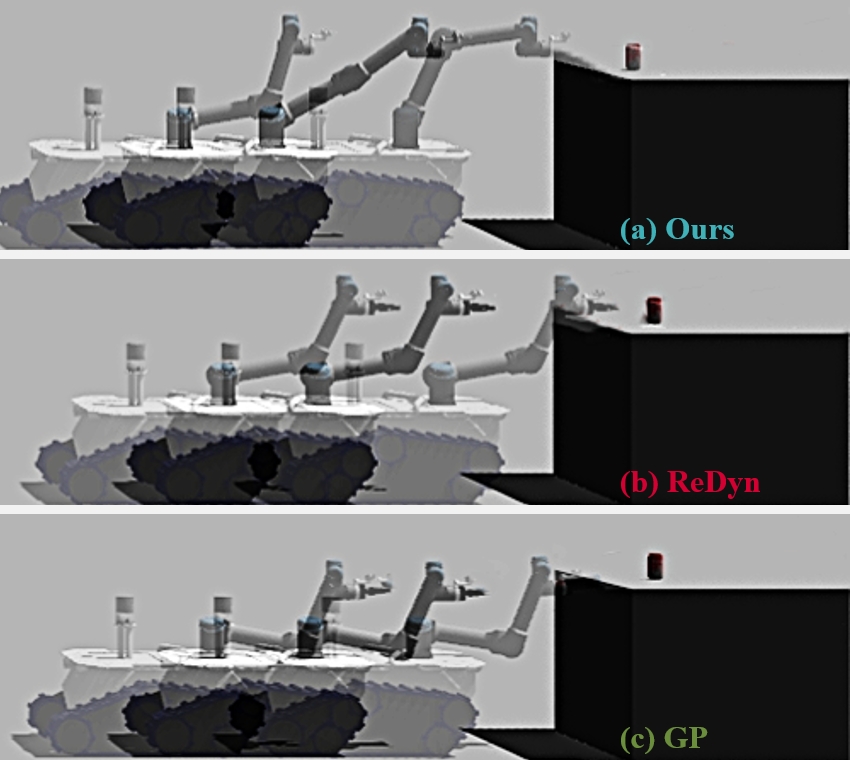}
	\captionsetup{font={small},justification=raggedright}
	\caption{The typical performance of different algorithms in grasping near obstacles.}
	\label{fig:07}
\end{figure}

The simulated scenario illustrated in Fig.~\ref{fig:06}(a) is designed to evaluate the obstacle-avoidance capability of various planning algorithms during a standard object grasping task. In all experiments, the robot's initial base path is a straight line from $(1.0\mathrm{m}, 0.0\mathrm{m}, 0.0\mathrm{m})$ to $(9.5\mathrm{m}, 0.0\mathrm{m}, 0.0\mathrm{m})$ with a step size of $0.1\mathrm{m}$. The manipulator's initial joint configuration path is set to a uniformly \textbf{elbow-down} posture. A solid platform with the grasping target (a canned beverage) is placed directly in front of the robot's initial position. Following the initial path without adaptation would result in a collision between the robot and the platform.

\begin{figure*}[t]
	\begin{minipage}{0.245\linewidth}
		\vspace{1pt}
		\subfloat[][Obstacle-aware grasping]{\includegraphics[width=1.0\linewidth]{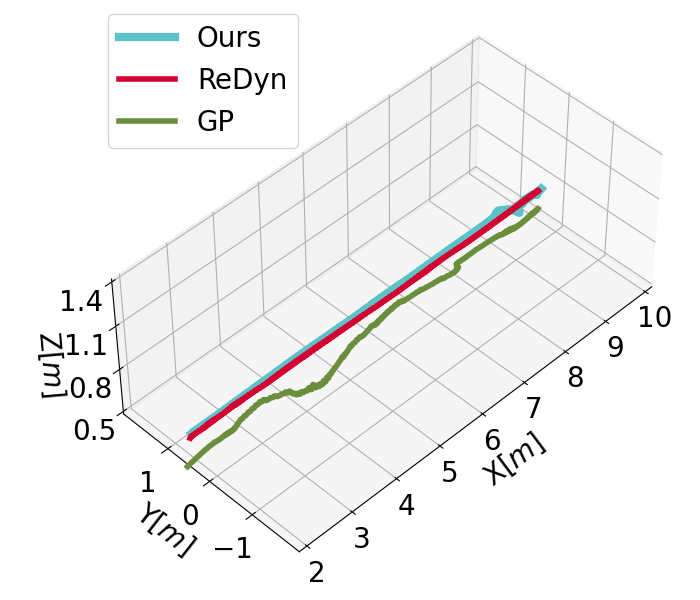}}
	\end{minipage}
	\begin{minipage}{0.245\linewidth}
		\vspace{1pt}
		\subfloat[][Grasping during locomotion]{\includegraphics[width=1.0\linewidth]{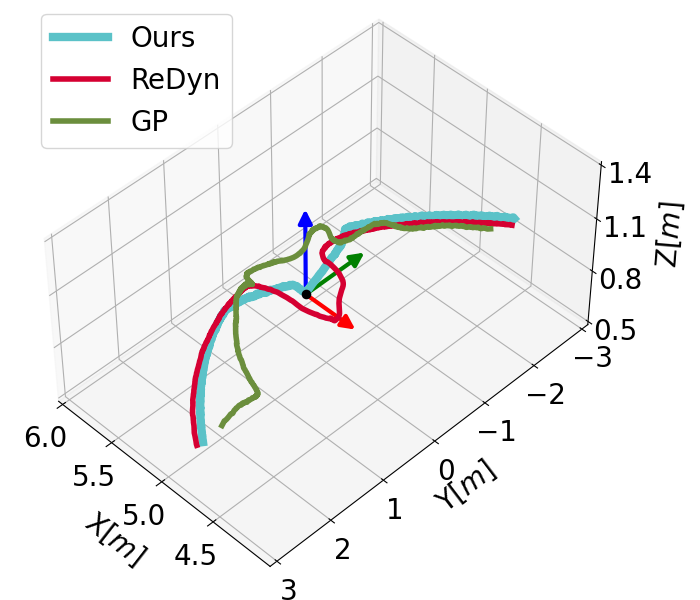}}
	\end{minipage}
	\begin{minipage}{0.245\linewidth}
		\vspace{1pt}
		\subfloat[][Inspection during locomotion]{\includegraphics[width=1.0\linewidth]{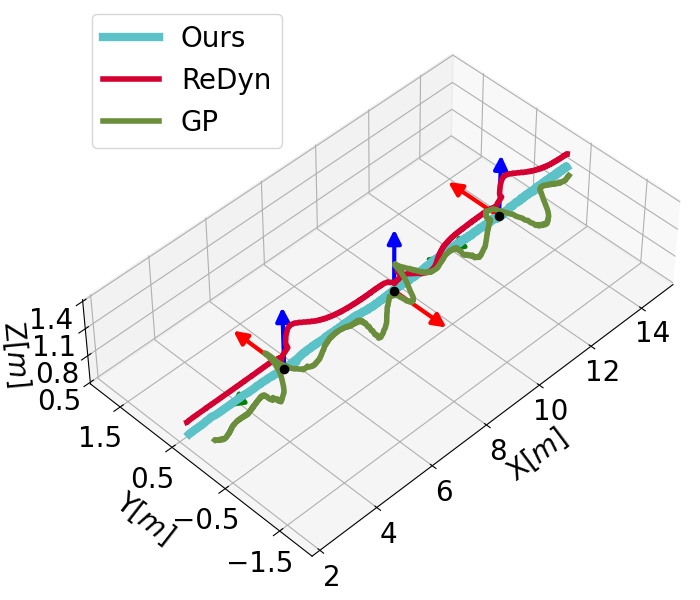}}
	\end{minipage}
	\begin{minipage}{0.245\linewidth}
		\vspace{1pt}
		\subfloat[][Payload transportation]{\includegraphics[width=1.0\linewidth]{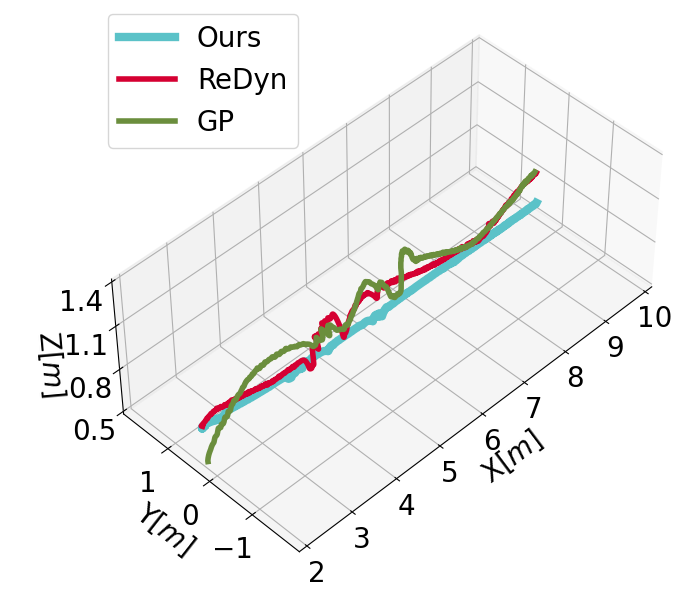}}
	\end{minipage}
	\captionsetup{font={small},justification=raggedright}
	\caption{The typical end-effector executed trajectories of all compared methods across the four simulated tasks. The coordinate frames in (b-c) indicates the optimal grasp/inspection pose, respectively.}
	\label{fig:08}
\end{figure*}

As shown in Fig.~\ref{fig:07}(a) and Fig.~\ref{fig:08}(a), the proposed framework generates feasible manipulator configurations considering obstacles around the target and applies a stability-aware interpolation strategy to achieve smooth configuration transitions, thereby enabling successful and collision-free grasping. In contrast, ReDyn only considers end-effector and base obstacle avoidance without accounting for the arm configuration, leading to collisions with the platform (Fig.~\ref{fig:07}(b)). Although GP incorporates manipulator collision costs into the objective function, poor initial solutions could prevent the iterative nonlinear optimizer from guiding the manipulator to transition from the elbow-down to elbow-up configuration across singularities, resulting in collision failure as well (shown in Fig.~\ref{fig:07}(c)).

\subsection{Grasping During Locomotion}

The simulation scenario depicted in Fig.~\ref{fig:06}(b) is used to evaluate the performance of different methods for manipulation while moving. In all experiments, the robot's initial base path is a circle with a radius of $5.5\mathrm{m}$ centered at $(0.0\mathrm{m}, 0.0\mathrm{m}, 0.0\mathrm{m})$, discretized with a step size of approximately $0.1\mathrm{m}$. The manipulator's initial joint configuration path is set to a uniformly \textbf{elbow-up} posture. The target object (a canned beverage) is placed at the edge of the platform. Its optimal grasp pose is indicated by a coordinate frame shown in Fig.~\ref{fig:08}(b).

\begin{figure}[h]
	\centering
	\includegraphics[width=0.49\textwidth]{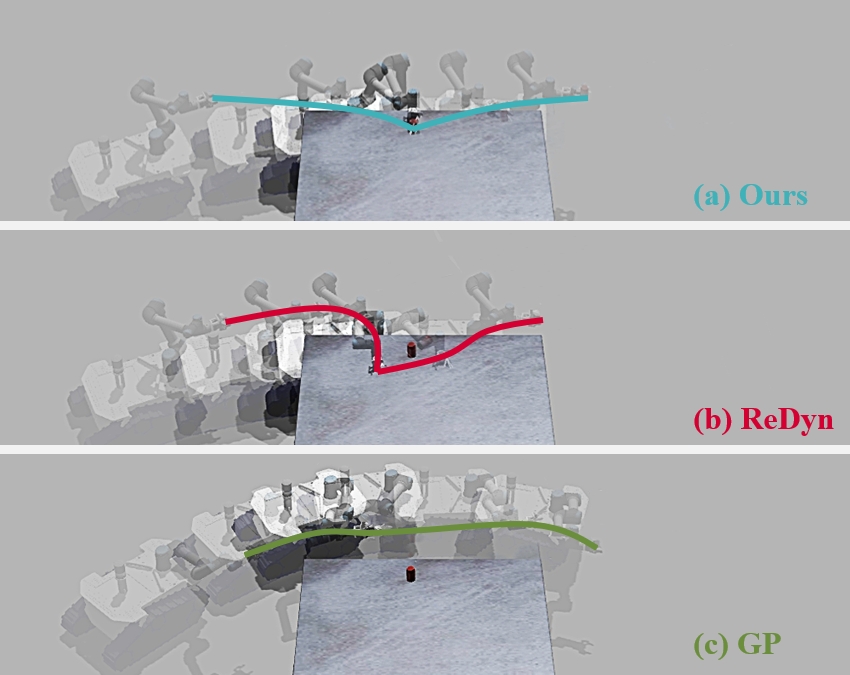}
	\captionsetup{font={small},justification=raggedright}
	\caption{The typical executed trajectories of the end-effector in grasping during locomotion.}
	\label{fig:09}
\end{figure}

As the executed trajectories illustrated in Fig.~\ref{fig:08}(b) and Fig.~\ref{fig:09}, the proposed framework incorporates hard constraints on the end-effector fixed-pose within the optimization, ensuring stability of the end-effector at the grasping point. With the IHC strategy, the framework effectively suppresses disturbances from base motion at the end-effector. In contrast, although GP enforces similar hard constraints on the end-effector fixed-pose, it exhibits significant deviations at higher base velocities. This is mainly due to the inherent non-convexity of its whole-body motion model, cost functions, and constraints, which complicates the SQP formulation, reduces planning efficiency, and can even lead to control overshoot. ReDyn exhibits degraded performance when the base velocity exceeds $0.6\mathrm{m/s}$, due to insufficient stabilization time and chassis velocity bias, leading to premature grasping and lower success rates. Its performance also shows strong dependence on the grasp pose, with suboptimal grasping outcomes observed when the grasping orientation is horizontal.

\subsection{Environmental Inspection During Locomotion}

The simulation scenario shown in Fig.~\ref{fig:06}(c) is designed to evaluate the hand-eye stability of the mobile manipulator during environmental inspection tasks. In all simulations, the initial base path is a straight line from $(0.0\mathrm{m}, 0.0\mathrm{m}, 0.0\mathrm{m})$ to $(15.0\mathrm{m}, 0.0\mathrm{m}, 0.0\mathrm{m})$ with a step size of $0.1\mathrm{m}$. The manipulator's initial joint configuration path is uniformly set to the \textbf{elbow-up} posture. Several key viewpoints towards both sides of the initial path are pre-specified (shown in Fig.~\ref{fig:08}(c)), and the robot is required to perform environmental inspection while maintaining end-effector stability.

As shown in Fig.~\ref{fig:08}(c) and Table~\ref{table:02}, the proposed framework generates an overall smooth end-effector trajectory, with only minor orientation adjustments at key viewpoints to ensure camera observation quality. In contrast, GP does not explicitly optimize the end-effector state and fails to suppress disturbances induced by base motion effectively. ReDyn exhibits noticeable overshoot during large end-effector pose adjustments, leading to insufficient hand-eye stability.

\begin{table}[h]
	\centering
	\captionsetup{font={small},justification=raggedright}
	\caption{Standard deviations of the end-effector linear acceleration $\sigma_{la}\mathrm{(m/s^2)}$, angular acceleration $\sigma_{aa}\mathrm{(rad/s^2)}$, and the maximum curvature $\kappa_{\max} (\mathrm{m^{-1}})$ of the executed end-effector trajectories for the compared methods performing rescue tasks with end-effector fixed-pose constraints.}
	\small
	\begin{threeparttable}
	\begin{tabular}{lcccccc}
		\toprule
				& \multicolumn{3}{c}{Grasp during locomotion}
				& \multicolumn{3}{c}{Inspect during locomotion} \\
				\cmidrule(lr){2-4}\cmidrule(lr){5-7}
				& $\sigma_{la}\downarrow$ & $\sigma_{aa}\downarrow$ & $\kappa_{\max}\downarrow$
				& $\sigma_{la}\downarrow$ & $\sigma_{aa}\downarrow$ & $\kappa_{\max}\downarrow$ \\
		\midrule
		Ours	& \textbf{1.090} & \textbf{1.658} & 142.2
				& 1.726 & \textbf{1.500} & \textbf{44.8} \\
		ReDyn	& 1.358 & 2.837 & 349.4
				& \textbf{1.555} & 1.642 & 133.2 \\
		GP		& 1.434 & 1.814 & \textbf{88.6}
				& 2.914 & 3.838 & 177.6 \\
		\bottomrule
	\end{tabular}
	\begin{tablenotes}
	\footnotesize
	\item $\downarrow$ indicates that a smaller value of this metric is better.
	\end{tablenotes}
	\end{threeparttable}
	\label{table:02}
\end{table}

\subsection{Transport of Force-Sensitive Loads on Rugged Terrain}

The simulation scenario depicted in Fig.~\ref{fig:06}(d) is used to evaluate the end-effector stability during the transport of force-sensitive payloads over rugged terrain. The initial base path is a straight line from $(0.0\mathrm{m}, 0.0\mathrm{m}, 0.0\mathrm{m})$ to $(10.0\mathrm{m}, 0.0\mathrm{m}, 0.0\mathrm{m})$, with a step size of $0.1\mathrm{m}$. The manipulator's initial joint configuration is uniformly set to the \textbf{elbow-up} posture. The environment contains pits and uneven surfaces, requiring the robot to maintain end-effector stability while maneuvering.

As shown in Fig.~\ref{fig:08}(d), both ReDyn and GP rely on whole-body control and struggle to effectively suppress disturbances from rugged terrain, resulting in visibly jagged end-effector trajectories. The average linear acceleration standard deviations of the end-effector are $3.422\mathrm{m/s^2}$ and $3.772\mathrm{m/s^2}$, respectively. In contrast, the proposed method produces a predominantly straight end-effector trajectory with minimal terrain influence, achieving an average linear acceleration standard deviation of $2.354\mathrm{m/s^2}$ during task execution.

\section{Experiments}
\label{experiments}

\begin{figure}[h]
    \centering
	\includegraphics[width=1.0\linewidth]{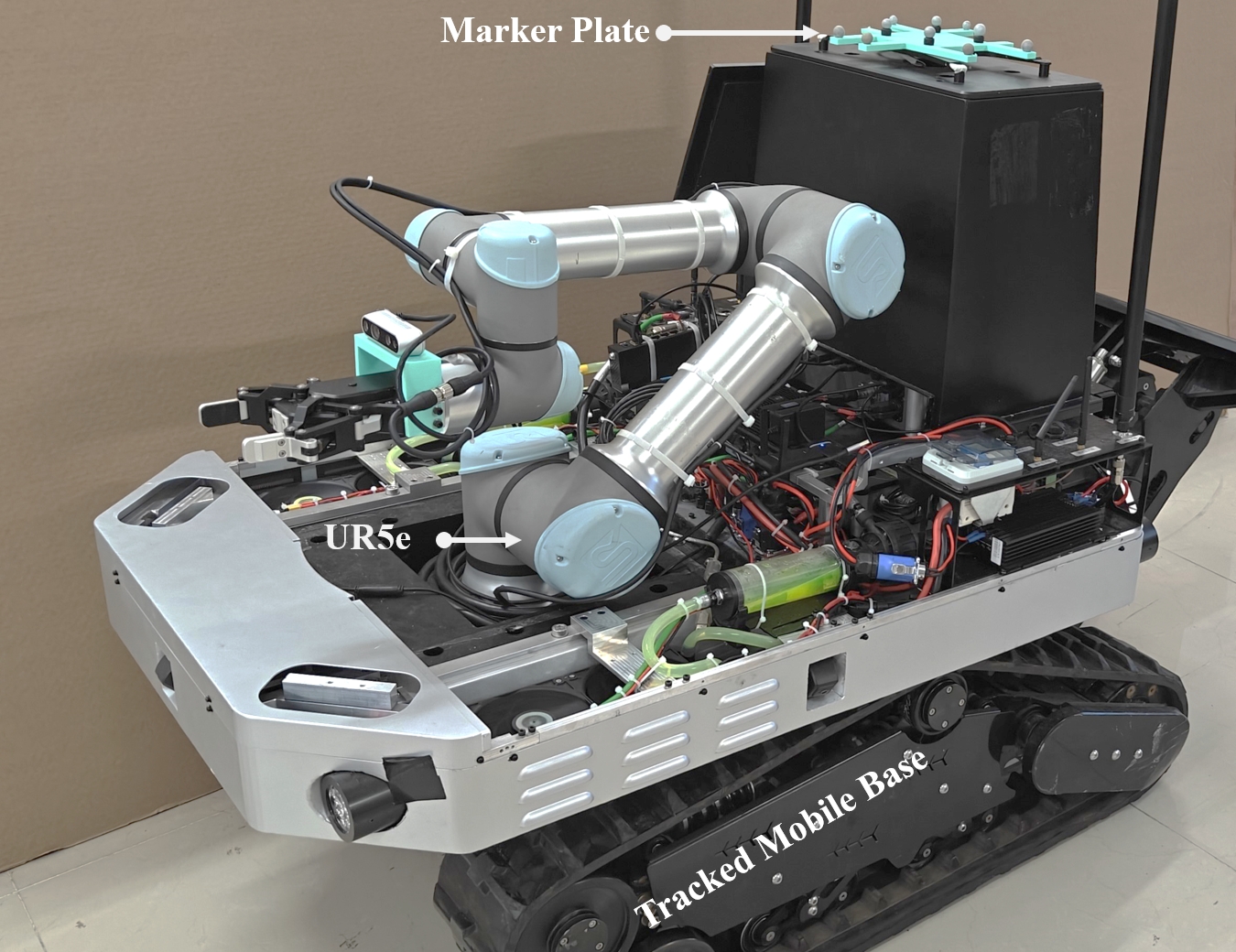}
    \captionsetup{font={small},justification=raggedright}
    \caption{Raigor: a self-developed tracked mobile manipulator.}
	\label{fig:10}
\end{figure}

This work utilizes a self-developed tracked mobile manipulator, \textbf{Raigor}, to validate the proposed method in the real-world scenarios. Raigor has a total weight of approximately $150\mathrm{kg}$, with a tracked chassis load capacity of $80\mathrm{kg}$, and is capable of stable locomotion across various complex and rugged terrains. The platform is equipped with a UR5e manipulator, featuring a maximum reach of $0.85\mathrm{m}$ and a maximum end-effector payload of $5\mathrm{kg}$ (see Fig.~\ref{fig:10}). The robot's end-effector is equipped with a two-finger gripper for grasping operations.

\begin{figure*}[t]
    \begin{minipage}{0.326\linewidth}
        \subfloat[][Grasping during locomotion]{\includegraphics[width=1.0\linewidth]{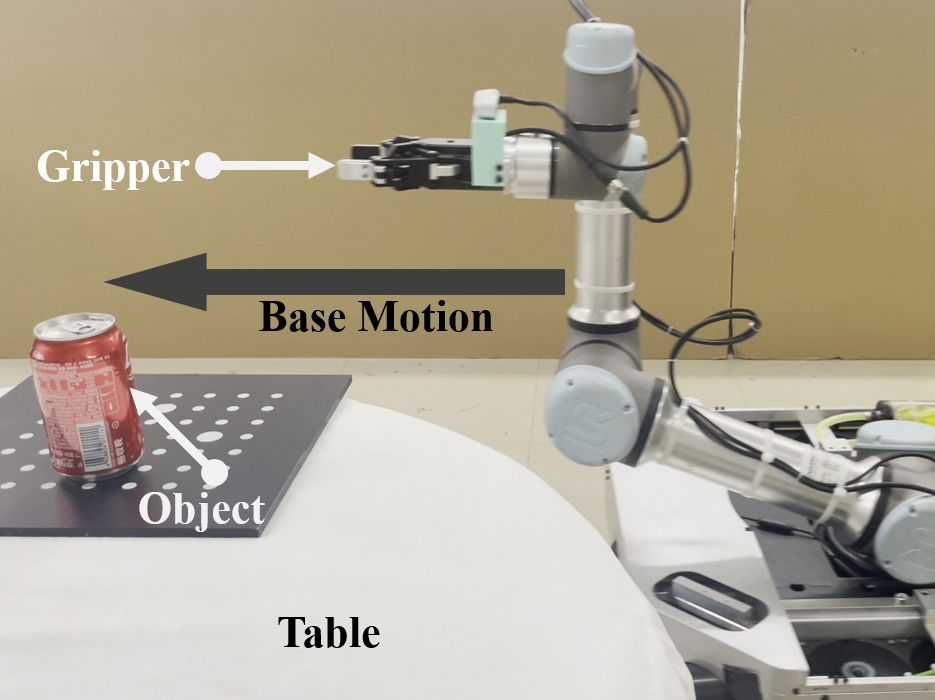}}
    \end{minipage}
    \begin{minipage}{0.6695\linewidth}
        \subfloat[][Object inspection on rugged terrain]{\includegraphics[width=1.0\linewidth]{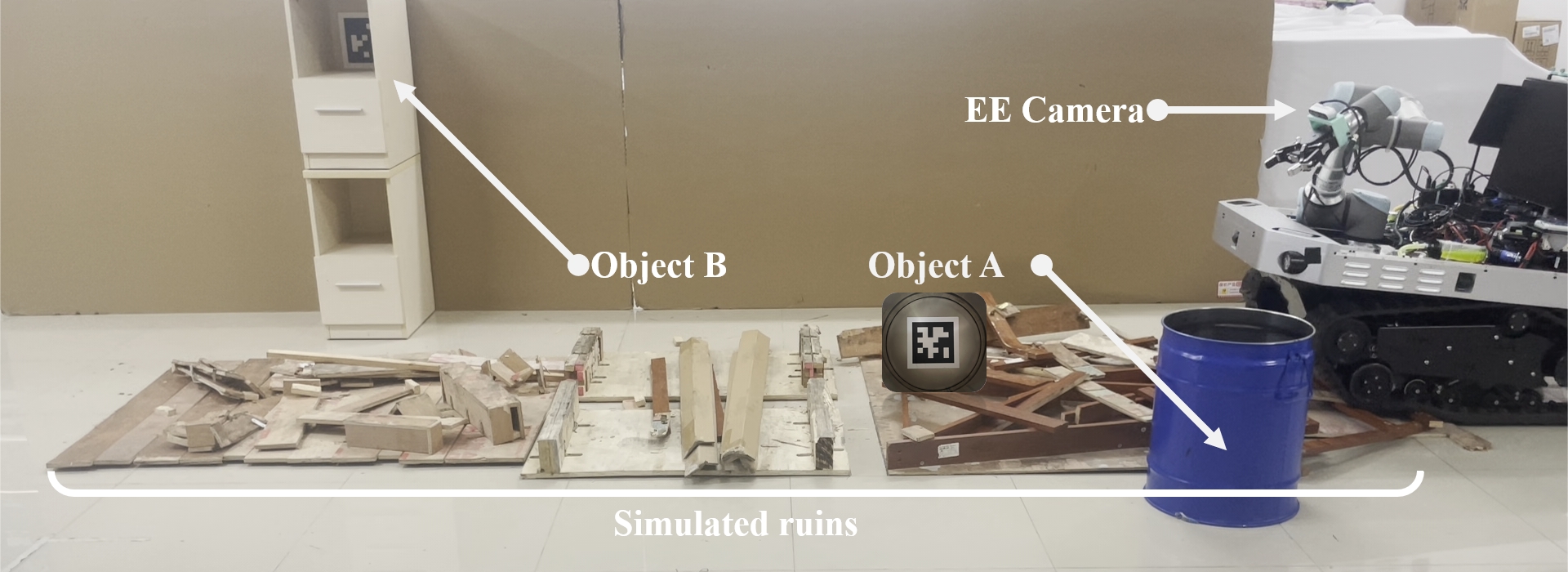}}
    \end{minipage}
    \captionsetup{font={small},justification=raggedright}
    \caption{Real-world experiment setups.}
	\label{fig:11}
\end{figure*}

During the experiments, multiple task scenarios were set to enable quantitative comparisons between different methods. Global localization of the robot was provided by a motion capture system, with the marker plate mounted on the robot's rear sensor frame. The environmental ESDF map was acquired using a handheld MID-360 LiDAR and subsequently constructed via HEATS~\cite{heats}. The onboard computer consists of an Intel i7-11300K processor with 8 cores running at 3.60~GHz. In addition to the proposed framework, ReDyn and GP were deployed on the real robot for comparative experiments.

\subsection{Grasping During Locomotion}

A series of object-grasping experiments was conducted to evaluate the proposed framework's effectiveness in maintaining end-effector stability and precise control during motion. The experimental setup is shown in Fig.~\ref{fig:11}(a), where the target objects were placed on a table adjacent to the robot's initial path. 

During the experiments, the robot's initial base path was defined as a straight line from $(-3.0\mathrm{m}, -0.6\mathrm{m}, 0.0\mathrm{m})$ to $(2.0\mathrm{m}, -0.6\mathrm{m}, 0.0\mathrm{m})$ with a step size of $0.1\mathrm{m}$. To comprehensively evaluate each method under varying task difficulties, two different grasping poses and corresponding initial manipulator configurations were tested. In the easy task setup, the manipulator's initial joint configuration was set to an \textbf{elbow-up} posture with a relatively accessible grasping pose. In the hard task setup, the robot was required to perform larger end-effector pose transformations from an initial \textbf{elbow-down} configuration that risked collision with the table, imposing stricter demands on both planning and control. 

\begin{figure*}[t]
	\begin{minipage}{0.33\linewidth}
		\vspace{1pt}
		\subfloat[][Grasping (Easy, $_{b}v_{max}^{x} = 0.3\mathrm{m/s}$)]{\includegraphics[width=1.0\linewidth]{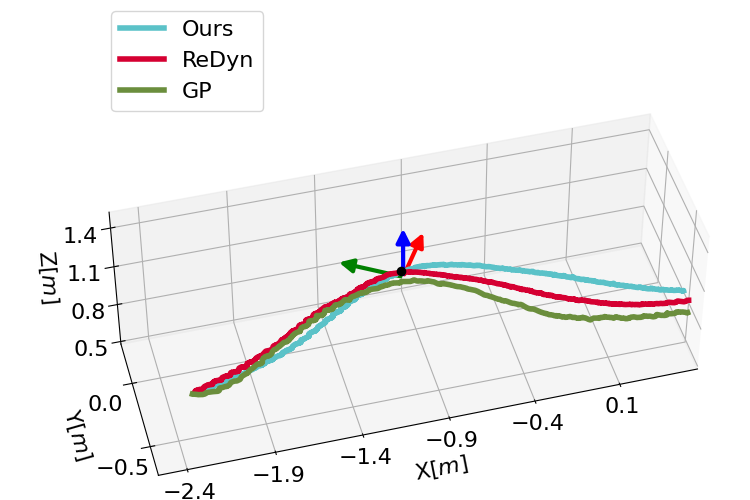}}
	\end{minipage}
	\begin{minipage}{0.33\linewidth}
		\vspace{1pt}
		\subfloat[][Grasping (Hard, $_{b}v_{max}^{x} = 0.3\mathrm{m/s}$)]{\includegraphics[width=1.0\linewidth]{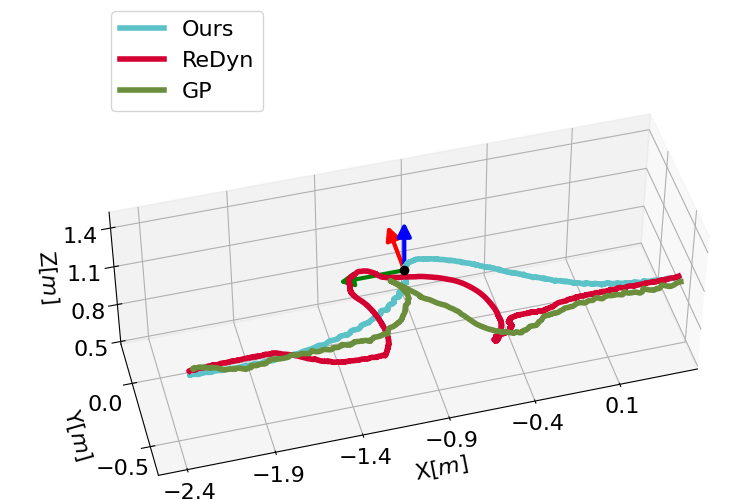}}
	\end{minipage}
	\begin{minipage}{0.33\linewidth}
		\vspace{1pt}
		\subfloat[][Object inspection on rugged terrain]{\includegraphics[width=1.0\linewidth]{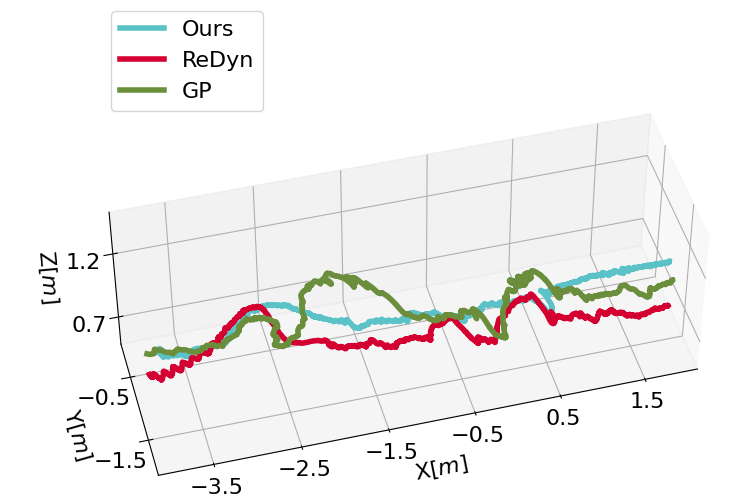}}
	\end{minipage}
	\captionsetup{font={small},justification=raggedright}
	\caption{The typical end-effector executed trajectories during the experiments. The coordinate frames in (a) and (b) indicate the optimal grasping poses.}
	\label{fig:12}
\end{figure*}

Additionally, experiments were repeated under three different maximum base velocities, $_{b}v_{max}^{x}$, namely $0.2\mathrm{m/s}$, $0.3\mathrm{m/s}$, and $0.5\mathrm{m/s}$, to assess the robot's performance under different dynamic conditions. When the base velocity was limited to $0.3\mathrm{m/s}$, the end-effector trajectories generated by each method are illustrated in Fig.~\ref{fig:12}(a)-(b).

\begin{figure*}[h]
	\begin{minipage}{0.01\linewidth}
		\vspace{3pt}
		\raggedright
		\rotatebox{90}{\fontsize{7}{7}\selectfont \textbf{Ours}}
	\end{minipage}
	\begin{minipage}{0.19\linewidth}
		\vspace{3pt}
		\centerline{\includegraphics[width=\textwidth]{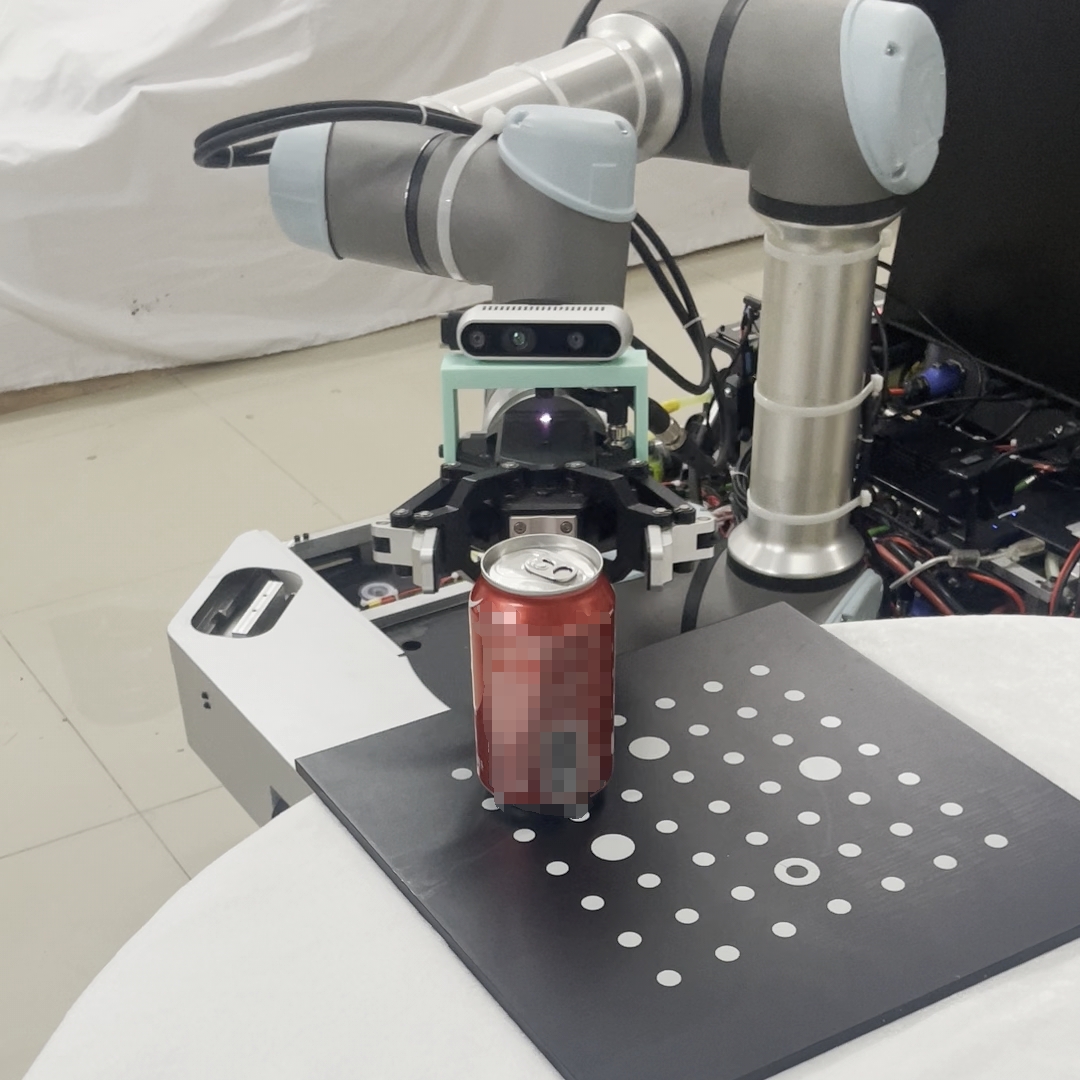}}
	\end{minipage}
	\begin{minipage}{0.19\linewidth}
		\vspace{3pt}
		\centerline{\includegraphics[width=\textwidth]{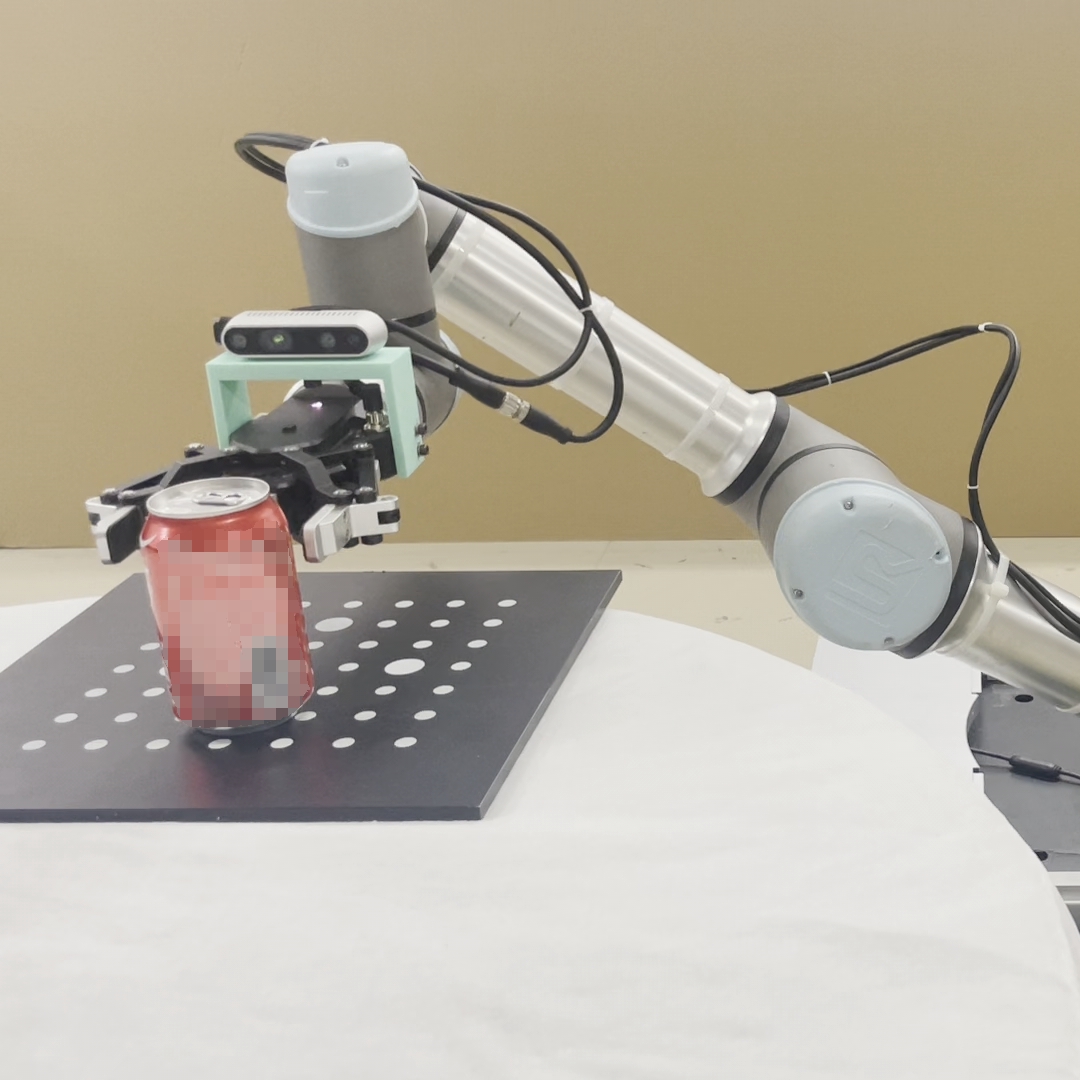}}
	\end{minipage}
	\begin{minipage}{0.57\linewidth}
		\vspace{3pt}
		\centerline{\includegraphics[width=\textwidth]{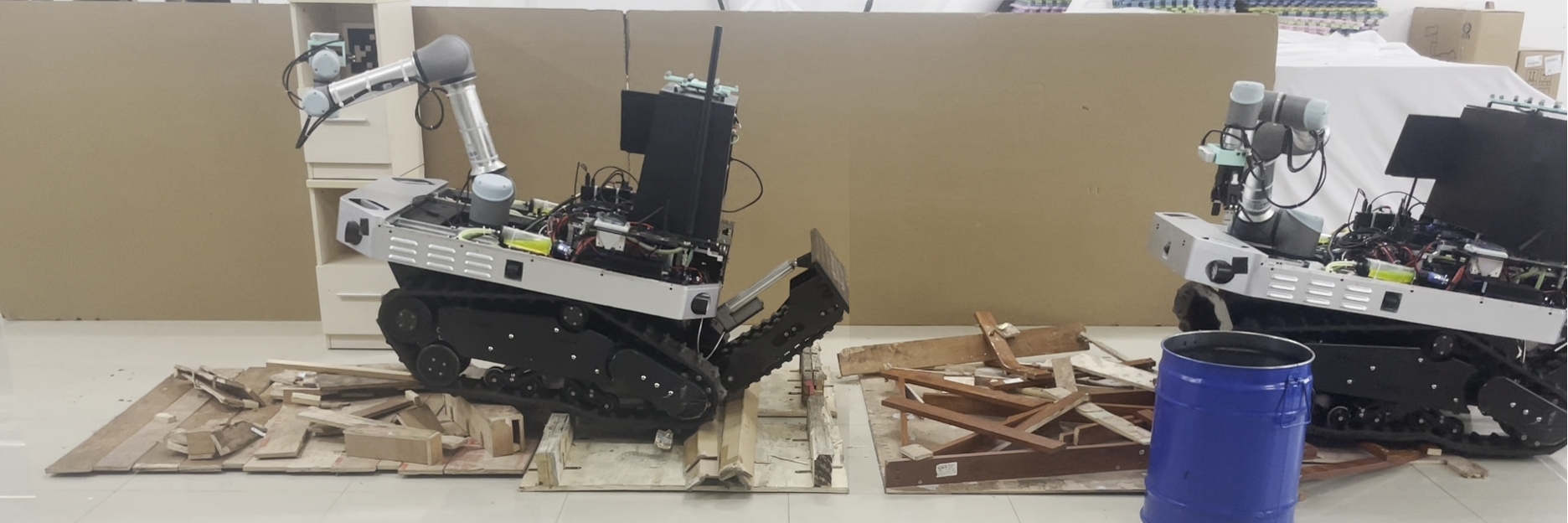}}
	\end{minipage}
	\\
	\begin{minipage}{0.01\linewidth}
		\vspace{3pt}
		\raggedright
		\rotatebox{90}{\fontsize{7}{7}\selectfont \textbf{ReDyn}}
	\end{minipage}
	\begin{minipage}{0.19\linewidth}
		\vspace{3pt}
		\centerline{\includegraphics[width=\textwidth]{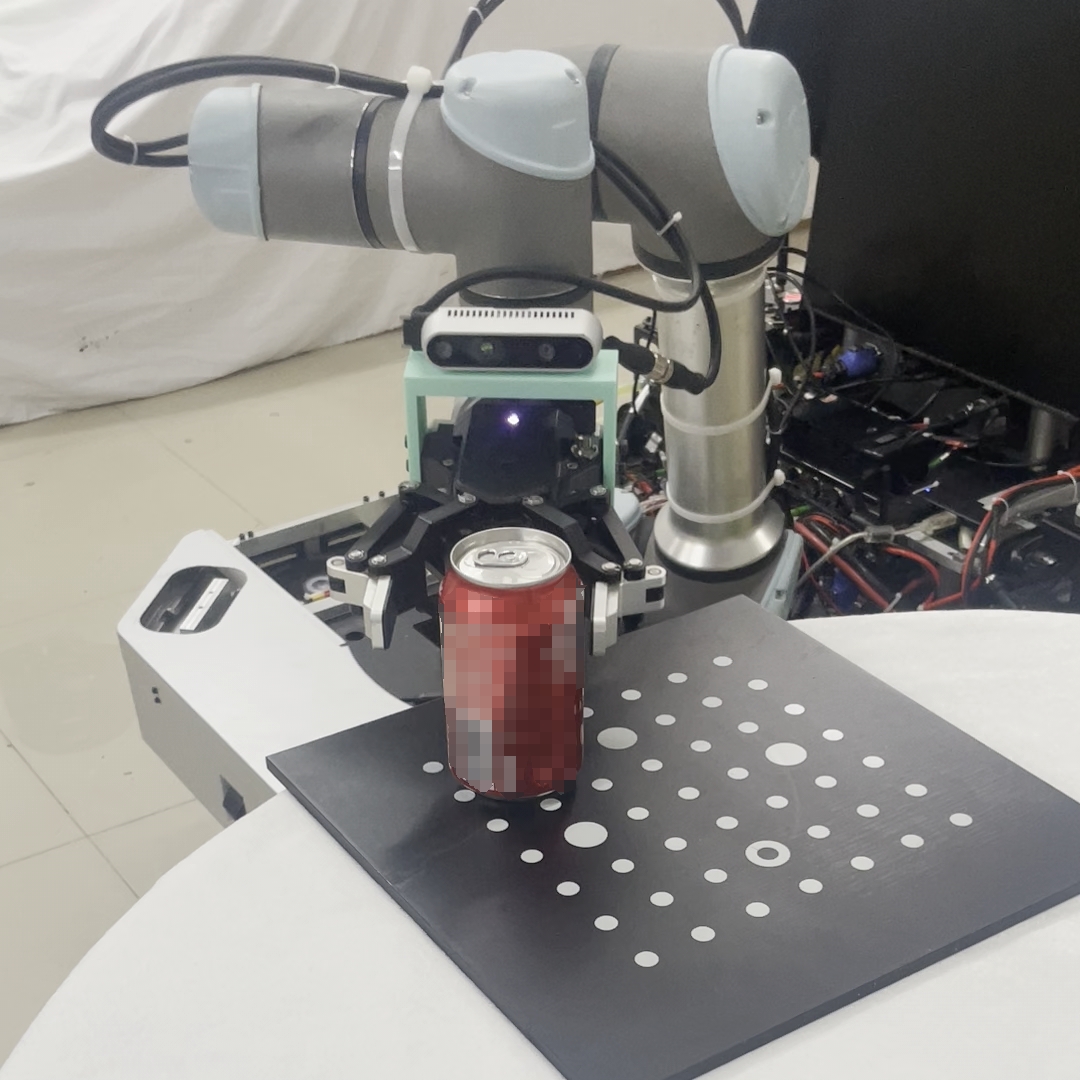}}
	\end{minipage}
	\begin{minipage}{0.19\linewidth}
		\vspace{3pt}
		\centerline{\includegraphics[width=\textwidth]{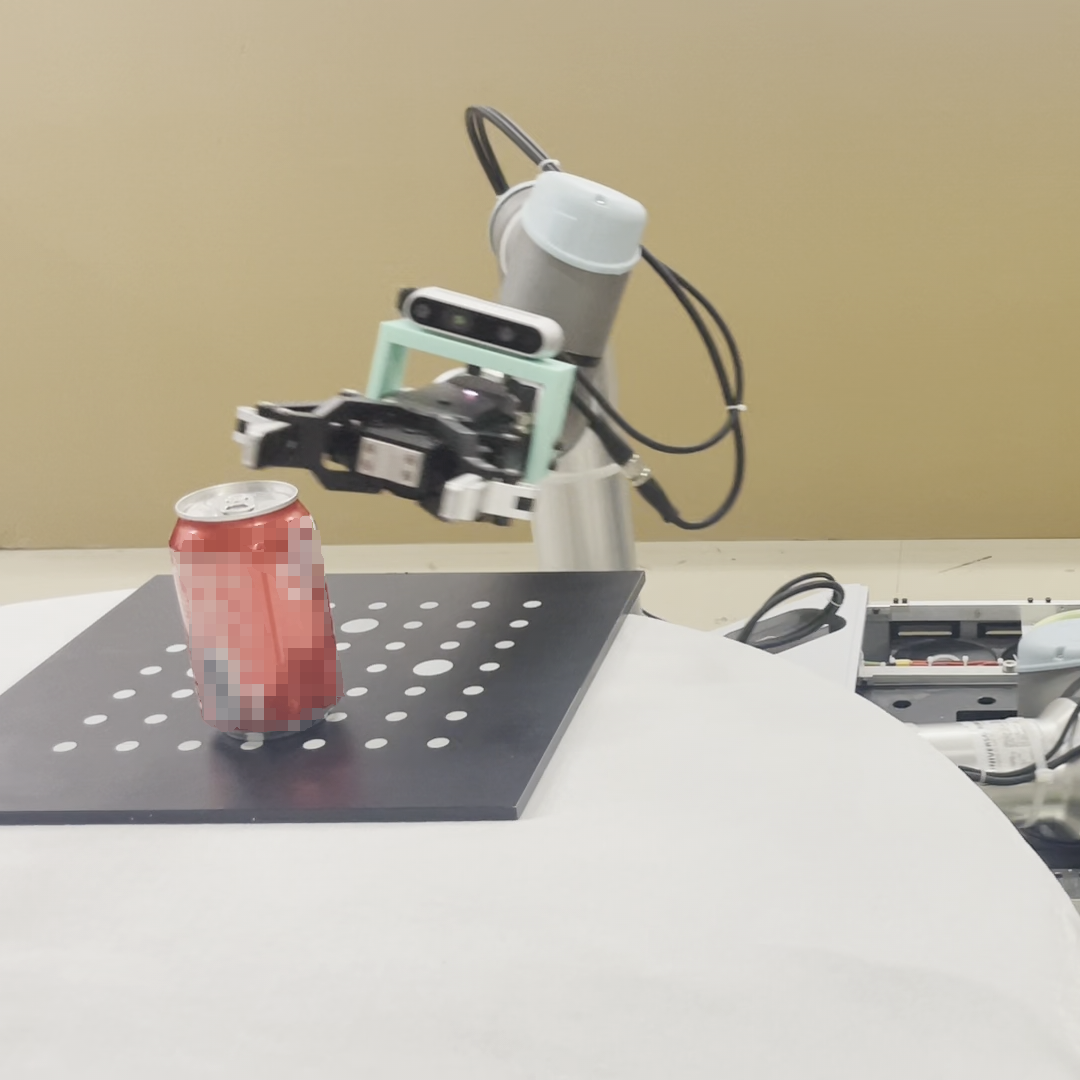}}
	\end{minipage}
	\begin{minipage}{0.57\linewidth}
		\vspace{3pt}
		\centerline{\includegraphics[width=\textwidth]{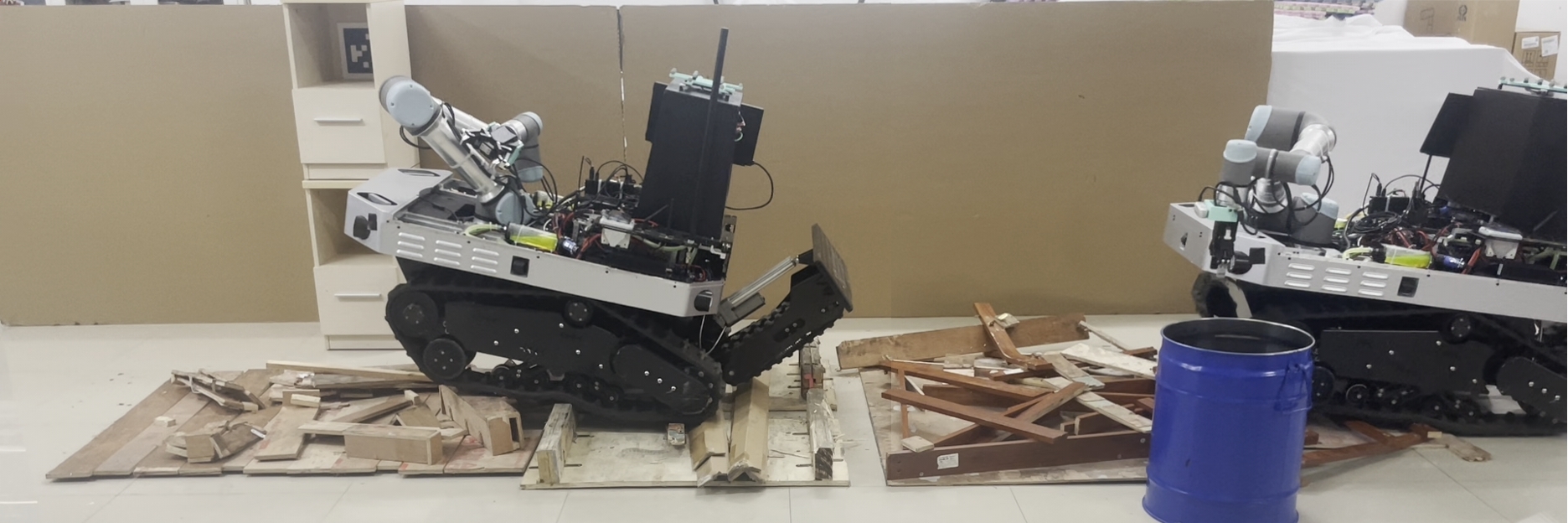}}
	\end{minipage}
	\\
	\begin{minipage}{0.01\linewidth}
		\vspace{3pt}
		\raggedright
		\rotatebox{90}{\fontsize{7}{7}\selectfont  \textbf{GP}}
	\end{minipage}
	\begin{minipage}{0.19\linewidth}
		\vspace{3pt}
		\captionsetup{justification=centering}
		\subfloat[][Grasping (Easy)]{\includegraphics[width=1.0\linewidth]{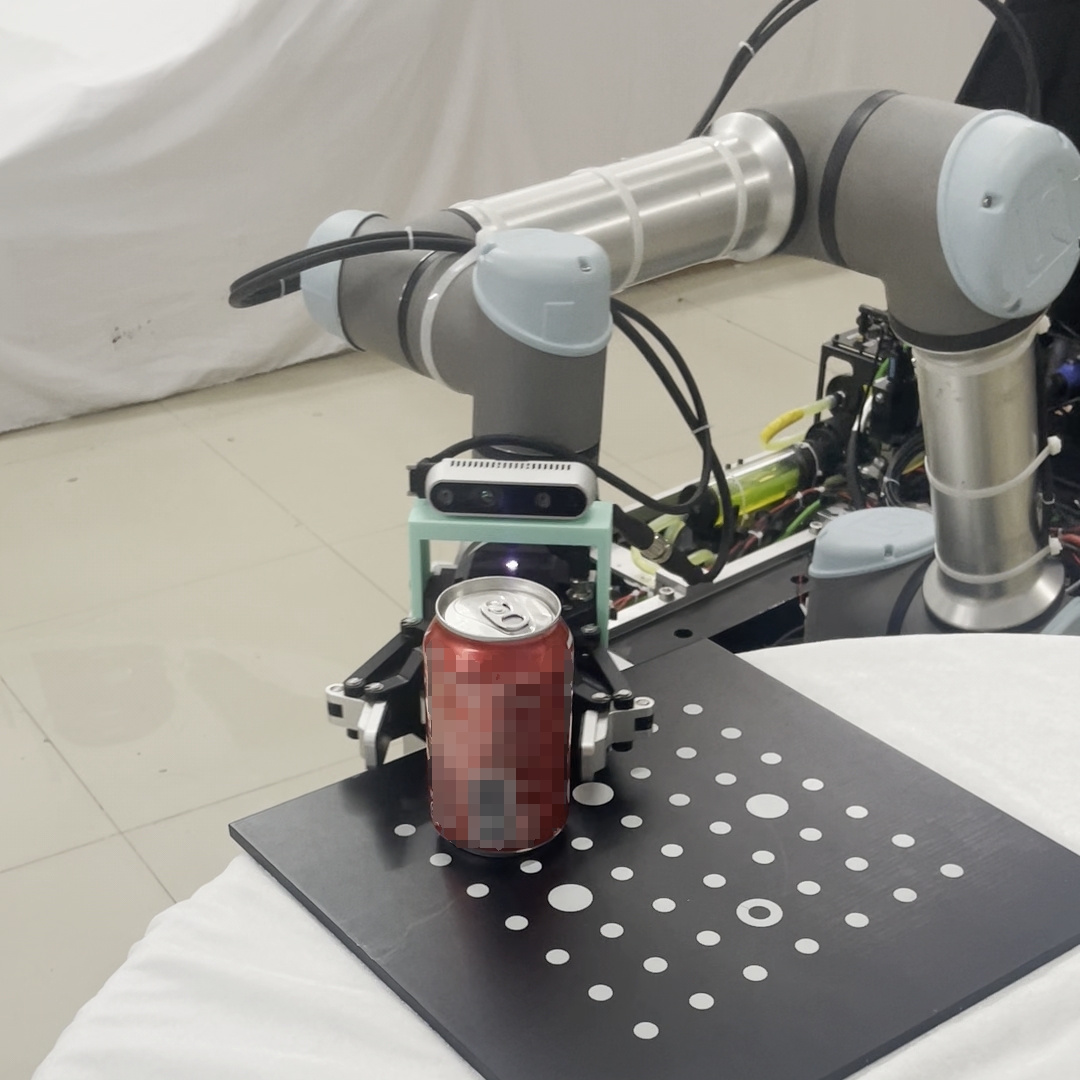}}
	\end{minipage}
	\begin{minipage}{0.19\linewidth}
		\vspace{3pt}
		\captionsetup{justification=centering}
		\subfloat[][Grasping (Hard)]{\includegraphics[width=1.0\linewidth]{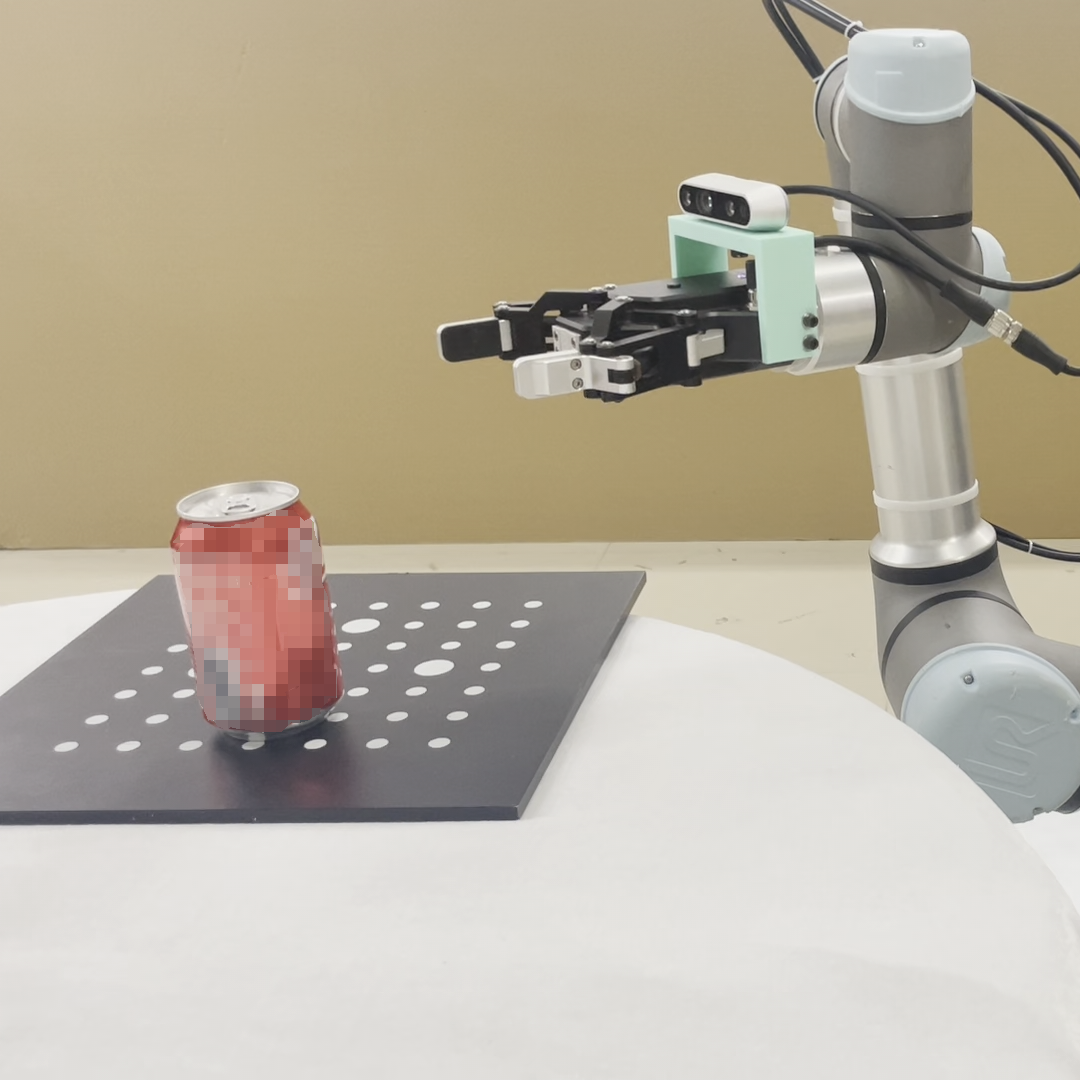}}
	\end{minipage}
	\begin{minipage}{0.57\linewidth}
		\vspace{3pt}
		\captionsetup{justification=centering}
		\subfloat[][Object inspection on rugged terrain]{\includegraphics[width=1.0\linewidth]{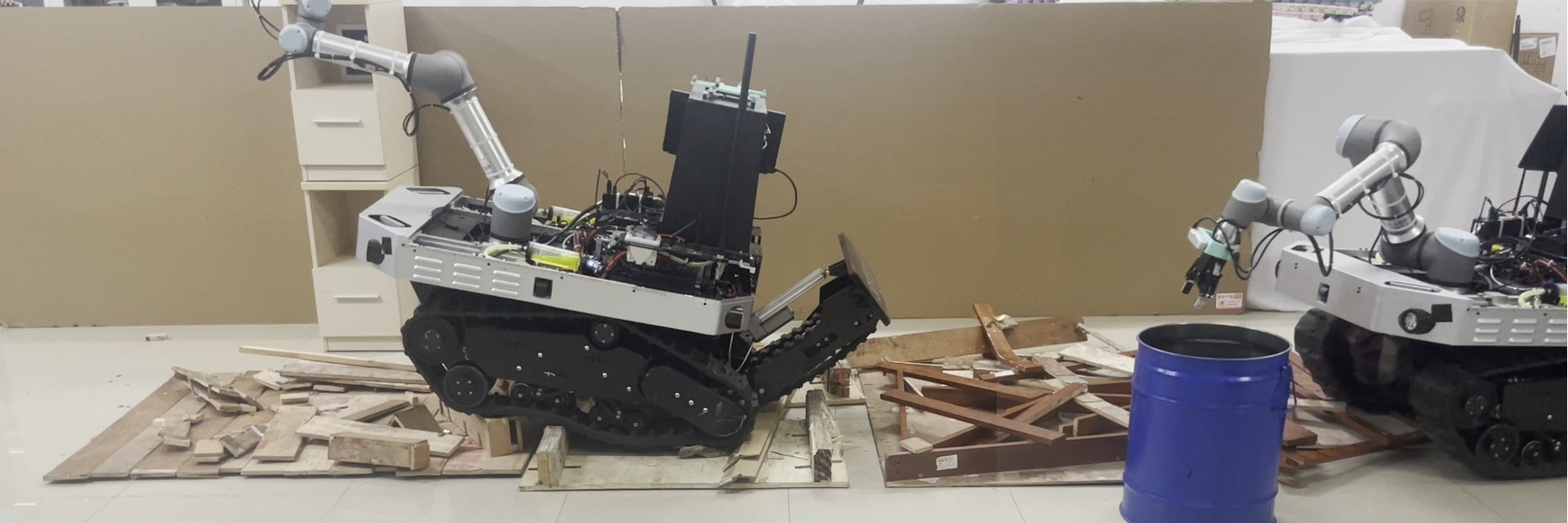}}
	\end{minipage}
	\captionsetup{font={small},justification=raggedright}
	\caption{The typical robot performance during the tasks: (a-b) the end-effector states in grasping during locomotion, captured either $0.1\mathrm{s}$ before gripper closure or at the moment when the end-effector is closest to the grasping pose; (c) the end-effector states near the objects during the object inspection task on rugged terrain.}
	\label{fig:13}
\end{figure*}

Under different conditions, each algorithm was independently tested five times. The results indicate that for simple grasping tasks under slow chassis motion, all methods maintain high success rates (see Table~\ref{table:03} and Fig.~\ref{fig:13}(a)). However, as the maximum chassis velocity increases, the success rates of ReDyn and GP decrease significantly. Specifically, ReDyn overly relies on inner-outer loop parameter tuning near the object, and because the HRQP controller tends to avoid low arm manipulability, the end-effector reaches the grasping pose prematurely at higher speeds, causing collisions. For GP, higher chassis velocities exacerbate unavoidable weight imbalances in the optimization problem, reducing end-effector control accuracy and compromising grasp stability.

\begin{table}[h]
	\centering
	\captionsetup{font={small},justification=raggedright}
	\caption{Success rates of the different algorithms under varying chassis velocity limitations $_{b}v_{max}^{x}(\mathrm{m/s})$ and task difficulty setups.}
	\small
	\begin{tabular}{lcccccc}
	\toprule
		& \multicolumn{3}{c}{Easy Scene} & \multicolumn{3}{c}{Hard Scene} \\
	\cmidrule(lr){2-4} \cmidrule(lr){5-7}
	$_{b}v_{max}^{x}$	& $\leq 0.2$ & $\leq 0.3$ & $\leq 0.5$ & $\leq 0.2$ & $\leq 0.3$ & $\leq 0.5$ \\
	\midrule
	Ours     & \textbf{100\%} & \textbf{100\%} & \textbf{100\%}  & \textbf{100\%} & \textbf{60\%} & \textbf{40\%} \\
	ReDyn     & \textbf{100\%} & 80\%           & 0\%            & 0\%            & 0\%           & 0\% \\
	GP & \textbf{100\%} & 40\%           & 20\%           & 0\%            & 0\%           & 0\% \\
	\bottomrule
	\end{tabular}
	\label{table:03}
\end{table}

For difficult grasping tasks, ReDyn only considers partial-body obstacle avoidance (i.e., the end-effector and base) and does not model the arm's collision constraints, resulting in multiple collisions with the table. Although GP incorporates obstacle costs into the optimization, it handles them rather directly and lacks modeling of manipulator configuration changes. Consequently, the robot cannot smoothly adjust its arm configuration in response to obstacles, and the combined effect of hard constraints and obstacle costs prevents successful path optimization, causing missed grasping poses (as shown in Fig.~\ref{fig:13}(b)). In contrast, the proposed framework maintains high grasping success rates even under higher chassis velocities and complex task conditions by enforcing end-effector fixed-pose constraints, smooth configuration interpolation, and isolated holistic control. The executed end-effector trajectories are smoother, and success rates are significantly higher than those of the compared methods.

\subsection{Object Inspection on Rugged Terrain}

A target inspection experiment was designed on rugged terrain to evaluate the robustness of the proposed framework in complex task scenarios. As shown in Fig.~\ref{fig:11}(b), two inspection targets were placed on either side of the initial base path: one at the bottom of a barrel and the other on the upper compartment of a cabinet. In this scenario, the robot relies solely on the hand-eye for the inspection through a confined observation window near the target. Although the Euclidean distance between the two observation points is short, the required pose changes are significant, challenging both smooth arm configuration transitions and end-effector stability. The robot's initial chassis path was a straight line from $(-5.0\mathrm{m}, -1.2\mathrm{m}, 0.0\mathrm{m})$ to $(2.0\mathrm{m}, -1.2\mathrm{m}, 0.0\mathrm{m})$ with a step size of $0.1\mathrm{m}$, and the arm's initial joint configuration path was set to the \textbf{elbow-up} configuration. The robot was required to traverse the rugged terrain while simultaneously performing target inspection. For safety, the chassis velocity was limited to $0.3\mathrm{m/s}$.

Experimental results show that the proposed framework effectively suppresses the end-effector suffered disturbances caused by the rugged terrain. Through end-effector motion smoothness costs and chassis-constrained interpolation, arm configuration transitions remain smooth, significantly reducing the probability of singular or unsafe configurations. In contrast, GP exhibits a more aggressive chassis motion and frequently go-and-stop near the observation points due to hard end-effector constraints. Moreover, the relatively slow iterative optimization and less consideration of chassis motion smoothness result in overshoot and oscillations in end-effector motion (see Fig.~\ref{fig:12}(c)). ReDyn frequently generates unsafe configurations due to rapid changes in desired end-effector poses, ultimately causing the arm to jam. This is primarily because ReDyn's QP optimization struggles to satisfy end-effector stability and feasible joint configurations simultaneously. Furthermore, in repeated experiments, ReDyn consistently failed to inspect the second target (see Fig.~\ref{fig:13}(c)). While GP successfully inspected both targets, its motion delays significantly increased the task completion time. Overall, the proposed framework balances end-effector stability and motion smoothness on rugged terrain, achieving the best trade-off between task efficiency and safety.

\subsection{Ablation of the Isolated Holistic Control Strategy}

A dynamic grasping experiment was designed to validate the effectiveness of the proposed IHC strategy with the F3B controller. During the experiment, the robot base was controlled to perform periodic lateral swings, while the isolated control strategy maintained the end-effector holding a force-sensitive payload (a wine glass of liquid) at a desired pose. The maximum rotational velocity of the base was set to $0.2\mathrm{rad/s}$, and the motion period was $10\mathrm{s}$. At the start of the experiment, the robot's end-effector held the wine glass horizontally.

\begin{figure}[h]
	\centering
	\includegraphics[width=1.0\linewidth]{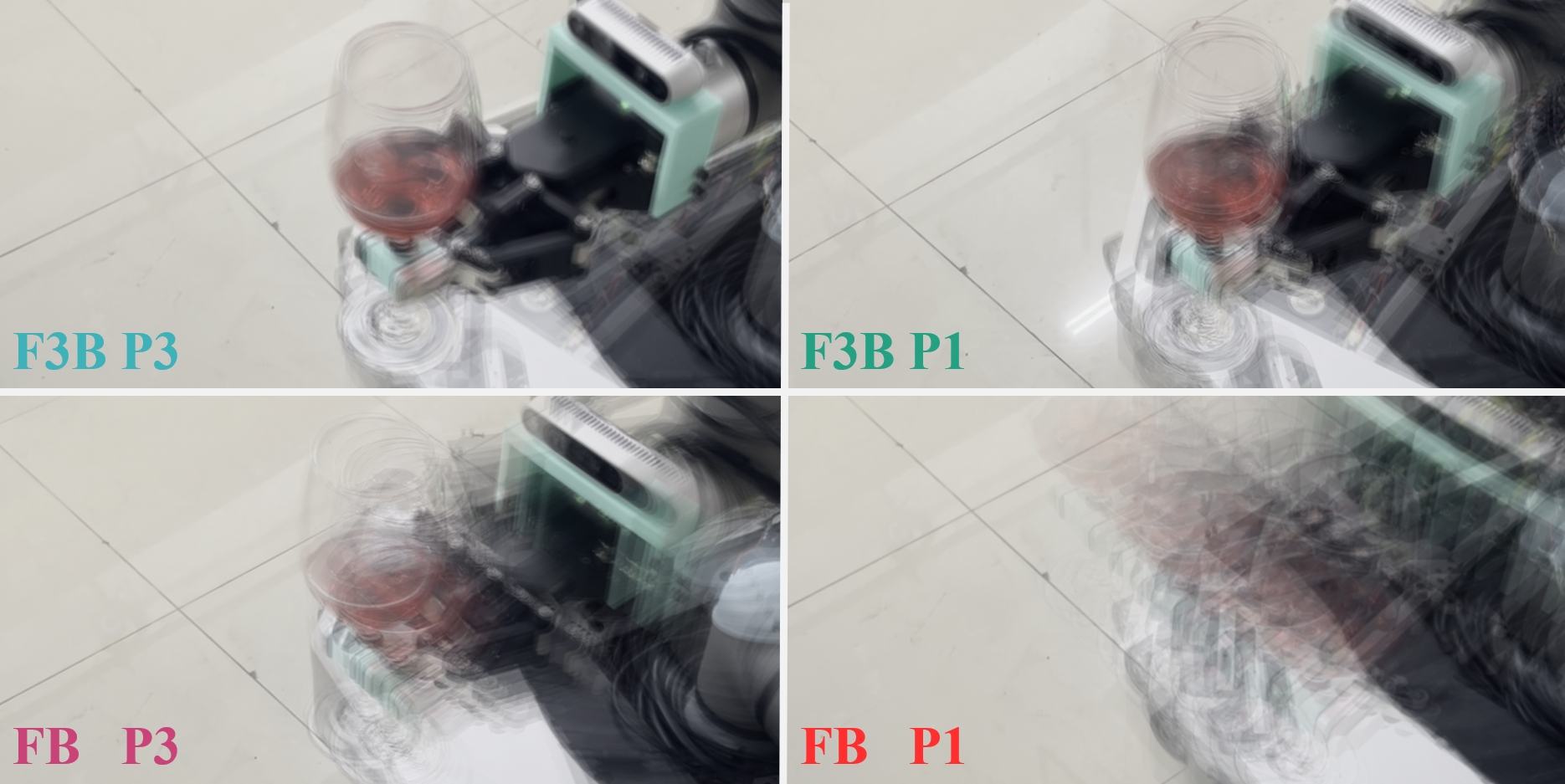}
	\captionsetup{font={small},justification=raggedright}
	\caption{The typical end-effector and payload motion traces in the ablation experiment ($15\mathrm{s}$ in total).}
	\label{fig:14}
\end{figure}

\begin{figure}[h]
	\centering
	\includegraphics[width=1.0\linewidth]{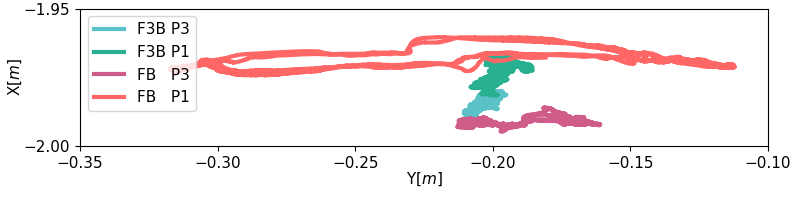}
	\captionsetup{font={small},justification=raggedright}
	\caption{The typical end-effector executed trajectories in the ablation experiment (top-down view).}
	\label{fig:15}
\end{figure}

\begin{figure}[h]
	\centering
	\includegraphics[width=1.0\linewidth]{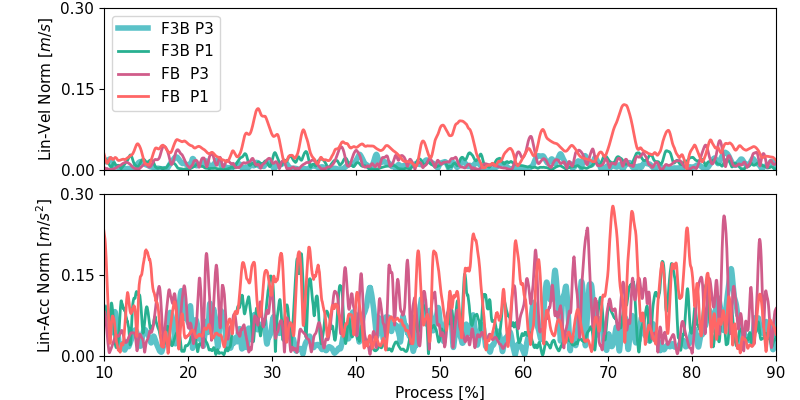}
	\captionsetup{font={small},justification=raggedright}
	\caption{The typical end-effector linear velocity and acceleration patterns in the ablation experiment.}
	\label{fig:16}
\end{figure}

In the ablation experiment, the F3B's proportional gain $K_p$ was chosen as the comparison parameter because it directly affects the closed-loop system's response speed. Specifically, $K_p\!=\!1.0$ and $K_p\!=\!3.0$ correspond to relatively weak and strong feedback strengths, respectively, representing typical control ranges. By comparing the robot's responses across different gains, the effectiveness of the proposed joint space feedforward projection in enhancing dynamic performance can be more intuitively evaluated. In addition to the experimental group using the F3B control strategy with a proportional gain of 3.0 (denoted as F3B-P3), three ablation baselines were tested: F3B-P1, which includes feedforward but has a proportional gain of 1.0; FB-P3, which excludes feedforward but has a proportional gain of 3.0; and FB-P1, which excludes feedforward and has a proportional gain of 1.0. Motion traces over $15\mathrm{s}$ were generated from video recordings (see Fig.~\ref{fig:14}), providing an intuitive visualization of end-effector stability under dynamic conditions.

\begin{table}[h]
	\centering
	\captionsetup{font={small},justification=raggedright}
 	\caption{Quantitative performance of different control strategies in real-world experiments: $p_{\mathrm{max}}\,(\mathrm{m})$ denotes the Euclidean distance between the farthest waypoints along the executed end-effector trajectory, $v_{\mathrm{mean}}\,(\mathrm{m/s})$ denotes the mean linear velocity of the end-effector in the global coordinate frame, $v_{\mathrm{max}}\,(\mathrm{m/s})$ denotes the maximum linear velocity of the end-effector in the global coordinate frame, $a_{\mathrm{mean}}\,(\mathrm{m/s^2})$ denotes the mean linear acceleration of the end-effector in the global coordinate frame, and $a_{\mathrm{max}}\,(\mathrm{m/s^2})$ denotes the maximum linear acceleration of the end-effector in the global coordinate frame.}
	\small
	\begin{threeparttable}
	\begin{tabular}{lccccc}
		\toprule
		 & $p_{\mathrm{max}}\downarrow$ & $v_{\mathrm{mean}}\downarrow$ & $v_{\mathrm{max}}\downarrow$ & $a_{\mathrm{mean}}\downarrow$ & $a_{\mathrm{max}}\downarrow$ \\ 
		 \midrule
		 F3B-P3 & \textbf{0.0169} & \textbf{0.0097} & \textbf{0.0326} & \textbf{0.0525} & \textbf{0.2382} \\ 
		 F3B-P1 & 0.0236 & 0.0120 & 0.0407 & 0.0596 & 0.3041 \\ 
		 FB-P3 & 0.0521 & 0.0147 & 0.0651 & 0.0747 & 0.5472 \\ 
		 FB-P1 & 0.2052 & 0.0424 & 0.1210 & 0.0932 & 0.3130 \\
		\bottomrule
	\end{tabular}
	\begin{tablenotes}
	\footnotesize
	\item $\downarrow$ indicates that a smaller value of this metric is better.
	\end{tablenotes}
	\end{threeparttable}
	\label{table:04}
\end{table}

The proposed method exhibits clear advantages in terms of maximum end-effector displacement error and reduced variations in velocity and acceleration (see Table~\ref{table:04}, Fig.~\ref{fig:15}, and Fig.~\ref{fig:16}). Among the evaluated configurations, the F3B-P3 setting achieves the best overall performance, effectively suppressing disturbances induced by chassis motion and producing the densest and clearest end-effector motion trajectories.

Under the same proportional gain, the proposed feedforward design significantly improves system responsiveness. In contrast, feedback-only controllers are constrained by hardware limitations in real-world experiments, which restrict the maximum allowable gains and consequently slow down the system response, leading to degraded tracking performance. These results demonstrate that the proposed isolated control strategy enables high-precision end-effector tracking while robustly rejecting chassis-induced disturbances, thereby enhancing overall operational stability.

\section{Conclusion}

This paper presents an end-effector stability-oriented planning and control framework for tracked mobile manipulators, targeting complex and diverse mobile manipulation tasks in disaster response scenarios. The proposed framework employs explicitly structured optimization variables to ease the definition of multiple nonlinear costs, enforce various hard constraints on end-effector states, and balance obstacle avoidance with manipulability. In addition, by adopting an isolated control strategy that integrates feedback regulation and feedforward compensation, the framework effectively suppresses disturbances from chassis motion, enhancing control responsiveness and end-effector stability. The framework has been validated through extensive simulations and real-world experiments across multiple rescue tasks. The results demonstrate that, compared to existing approaches, the proposed method not only ensures safe mobility in complex terrains but also significantly improves end-effector manipulability and stability. Future work will focus on incorporating learning-based perception-planning-control integration to overcome model inaccuracies and limited coordination among control modules under complex disturbances, advancing end-to-end data-driven regulation and control frameworks. 

\bibliographystyle{IEEEtran}
\bibliography{emma_ref}{}

\balance
\vfill

\end{document}